\documentclass[lettersize,journal]{IEEEtran}
\usepackage{amsmath,amsfonts}
\usepackage{algorithmic}
\usepackage{algorithm}
\usepackage{array}
\usepackage[caption=false,font=normalsize,labelfont=sf,textfont=sf]{subfig}
\usepackage{textcomp}
\usepackage{stfloats}
\usepackage{url}
\usepackage{booktabs} 
\usepackage{verbatim}
\usepackage{graphicx}
\usepackage{xcolor}
\usepackage[table]{xcolor}
\usepackage{tabularx} 
\usepackage{cite}
\usepackage{placeins} 

\begin{document}

\title{Graph-to-Vision: Benchmarking Multi-Graph Visual Reasoning with Vision-Language Models}

\author{
Qihang Ai, Ruizhou Li, Menghui Wang, and Haiyun Jiang%
\thanks{Qihang Ai is with Nanyang Technological University, Singapore (e-mail: qihang005@e.ntu.edu.sg).}
\thanks{Ruizhou Li is with Shandong University, China (e-mail: 202300820010@mail.sdu.edu.cn).}
\thanks{Menghui Wang is with Fudan University, China (e-mail: wangmh19@fudan.edu.cn).}
\thanks{Haiyun Jiang is with Shanghai Jiao Tong University, China (corresponding author, e-mail: haiyunjiangnlp@gmail.com).}
}

\maketitle

\begin{abstract}
Recent advances in Vision-Language Models (VLMs) have shown promising capabilities in interpreting visualized graph data, offering a perspective on graph-structured reasoning that complements traditional Graph Neural Networks (GNNs) by dispensing with explicit graph construction. 
However, existing studies focus primarily on single-graph reasoning, leaving the critical challenge of multi-graph visual reasoning (MGVR) underexplored. 
In this work, we introduce the first comprehensive benchmark designed to evaluate and enhance the MGVR abilities of VLMs. 
Our benchmark covers four common graph types—knowledge graphs, flowcharts, mind maps, and route maps—and supports both homogeneous and heterogeneous graph groupings with tasks of increasing complexity. 
We evaluate several state-of-the-art VLMs under a multi-dimensional scoring framework that assesses graph parsing, reasoning consistency, and instruction-following accuracy. 
Additionally, we fine-tune multiple open-source models and observe improvements in 17 of the 18 model--dimension combinations, confirming the effectiveness of our dataset. 
This work provides a principled step toward advancing MGVR and reveals new opportunities for cross-modal graph intelligence.
\end{abstract}


\section{Introduction}
\label{section1}
\IEEEPARstart{G}raphs are fundamental for modeling complex relationships and are widely used in domains such as knowledge representation, social networks, and recommendation systems \cite{wu2022graph,11194129,10948338}. With the rise of deep learning, there is growing interest in reasoning over multiple graphs to support tasks like knowledge integration and complex decision-making.

Although Graph Neural Networks (GNNs) have shown strong performance in various graph-based tasks \cite{zhou2021graphneuralnetworksreview,10912770,10074111}, they face notable challenges in multi-graph settings—particularly with heterogeneous graph structures—due to scalability limitations and poor generalization \cite{wu2023discoveringexplainingrepresentationbottleneck}.
In practice, however, many graphs are available only as rendered diagrams, from which an explicit structural representation must first be recovered. Our study therefore focuses on this visual setting and treats VLM-based reasoning as complementary to structure-based learning on explicit graphs.

In parallel, Vision-Language Models (VLMs) \cite{chen2020uniteruniversalimagetextrepresentation, qi2025vision}, which combine Transformer-based encoders for text and images, have demonstrated promising cross-modal reasoning abilities. Recent work suggests that rendering graphs as images and feeding them into VLMs allows better generalization across diverse structures \cite{zou2024vgbench}. Building on this, DeepSeek-OCR~\cite{wei2025deepseek} introduces an LLM-centric paradigm for Contexts Optical Compression. By compressing dense optical information into a minimal set of visual tokens, this approach significantly enhances the efficiency and accuracy of high-resolution document understanding and complex layout interpretation.

However, most existing studies focus on single-graph reasoning. The ability to jointly interpret and reason across multiple graphs—critical for tasks like multi-source alignment or integrative analysis—remains underexplored. We refer to this task as multi-graph visual reasoning (MGVR). To address it, we introduce the first benchmark designed specifically for MGVR with VLMs. It covers four common graph types (flowcharts, knowledge graphs, mind maps, and route maps) and includes both homogeneous and heterogeneous groupings with progressively difficult tasks.

We explicitly distinguish MGVR from the broader notion of multi-figure or multi-image reasoning studied in the general VLM literature. The latter typically involves natural photographs, charts, or documents, where the primary challenge lies in perceptual alignment or content-level comparison across images. In contrast, MGVR requires models to parse the explicit relational and topological structures embedded in each graph image—including nodes, edges, hierarchies, and paths—and integrate these structured semantics across multiple graphs. This structural, relation-centric requirement is the core distinction that motivates our benchmark and separates it from general multi-image VQA settings.

We propose a multi-dimensional evaluation framework assessing graph parsing, instruction-following, and reasoning consistency. Using this benchmark, we evaluate several state-of-the-art VLMs and fine-tune open-source models, observing improved reasoning capabilities in 17 of the 18 model--dimension combinations measured.
Despite these contributions, our fine-tuning is currently limited to lightweight models due to the high computational cost of large-scale VLMs, restricting scalability analysis and broader applicability.
Our main contributions are as follows:
\begin{enumerate}
    \item We introduce the first comprehensive benchmark for evaluating and improving the MGVR abilities of VLMs.
    \item We systematically evaluate several state-of-the-art VLMs on our benchmark using a dedicated multi-dimensional framework designed for MGVR.
    \item We fine-tune multiple open-source VLMs on our benchmark and observe improved MGVR performance in 17 of the 18 model--dimension combinations measured.
\end{enumerate}

\section{Related Work}

Recent work has increasingly explored Vision-Language Models (VLMs) for graph reasoning, especially through visual modalities. Image-based benchmarks such as GRAPHTMI \cite{das2023modality}, VisionGraph \cite{li2024visiongraph}, and VGBench \cite{zou2024vgbench} demonstrate that visual formats often outperform text for structured reasoning. Diagram-oriented datasets like NovaChart \cite{hu2024novachart} and DiagramQG \cite{zhang2024diagramqg} further extend this direction to broader reasoning tasks. Despite these advances, recent studies on charts—a structured form of visual graphs—highlight that VLMs remain sensitive to visual perturbations and struggle with complex reasoning \cite{mukhopadhyay2024unraveling}.

To address such limitations, structured visual priors have been incorporated. GITA \cite{wei2024gita} leverages layout-aware visual graphs, and LLaVA-SG \cite{wang2025llava} introduces scene graph intermediates with message passing for relation-aware parsing. In optimization domains, Bridging \cite{zhao2025bridging} exploit graph-based visual cues for improved performance without parameter tuning.

Separately, a line of work studies reasoning over multiple images. MileBench evaluates multimodal models in long-context, multi-image settings \cite{song2024milebench}; MMIU covers diverse relationships and tasks involving multiple images \cite{meng2025mmiu}; Mantis develops interleaved multi-image instruction tuning for skills such as co-reference, comparison, and temporal reasoning \cite{jiang2024mantis}; and NLVR2 requires determining whether a statement is valid for a pair of natural photographs \cite{suhr2019nlvr2}. These works primarily compare or aggregate content across natural images and do not specifically target explicit node-edge semantics. In contrast, our benchmark requires cross-graph structural alignment and relational inference over nodes, edges, hierarchies, and paths.

While progress in single-graph reasoning and general multi-image reasoning is significant, MGVR remains largely unaddressed. Existing benchmarks lack mechanisms for evaluating cross-graph integration of explicit relational and topological structures. Our work fills this gap by introducing a dedicated benchmark for MGVR and evaluating modern VLMs under a multi-dimensional framework.

\section{Dataset}
\label{section3}
\subsection{Overview}
\begin{figure}[h]
    \centering
    \includegraphics[width=1.0\linewidth]{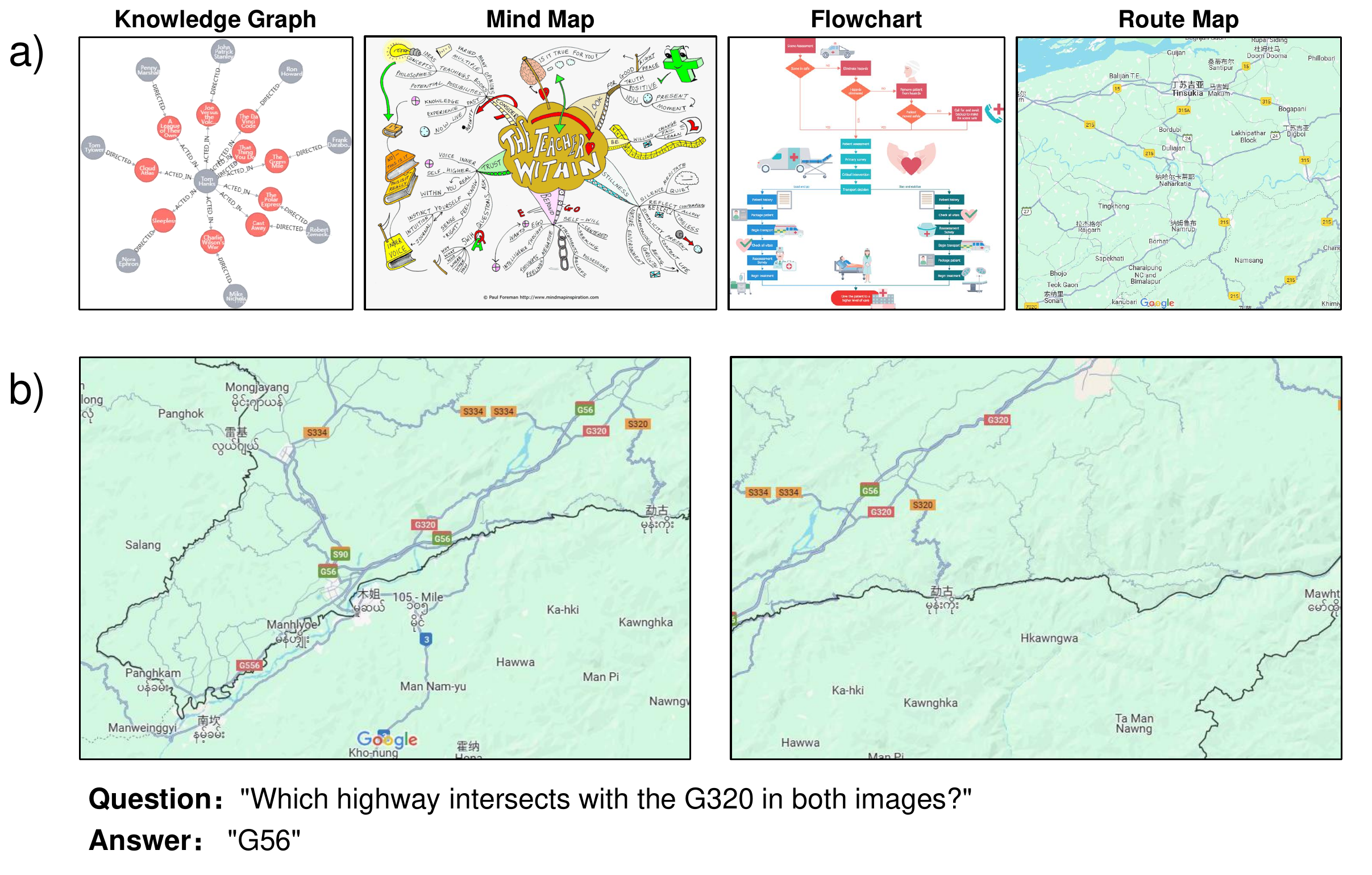}
    \caption{(a) Examples of the four types of graphs included in our benchmark: knowledge graphs, flowcharts, mind maps, and route maps. (b) An example sample from our benchmark, consisting of a set of related graphs, a corresponding instruction, and its reference answer.}
    \label{figure1,2}
\end{figure}

We introduce a benchmark specifically designed to evaluate the MGVR capabilities of Vision-Language Models (VLMs).
As illustrated in Figure \ref{figure1,2} (a), the benchmark includes four types of graph images—knowledge graphs, flowcharts, mind maps, and route maps—which reflect common structures in real-world reasoning tasks.
Each data sample in our benchmark consists of a set of interrelated graph images (e.g., graphs with shared themes, overlapping nodes, or logically connected content), a natural language instruction, and a ground-truth response.
An example of such a sample is provided in Figure \ref{figure1,2} (b).

To facilitate systematic evaluation, the image sets are organized into two categories:
(1) Homogeneous-type groups, where all graphs belong to the same category, and
(2) Heterogeneous-type groups, where graphs span different categories.
Each instruction is crafted to require reasoning across multiple graphs in the set, thereby assessing a model’s ability to jointly interpret and integrate graph-structured information.
All instruction-response pairs are initially generated by GPT-4o \cite{openai2024gpt4o}, followed by rigorous human verification, filtering, and refinement to ensure quality, clarity, and consistency.

\begin{figure}
    \centering
    \includegraphics[width=0.9\linewidth]{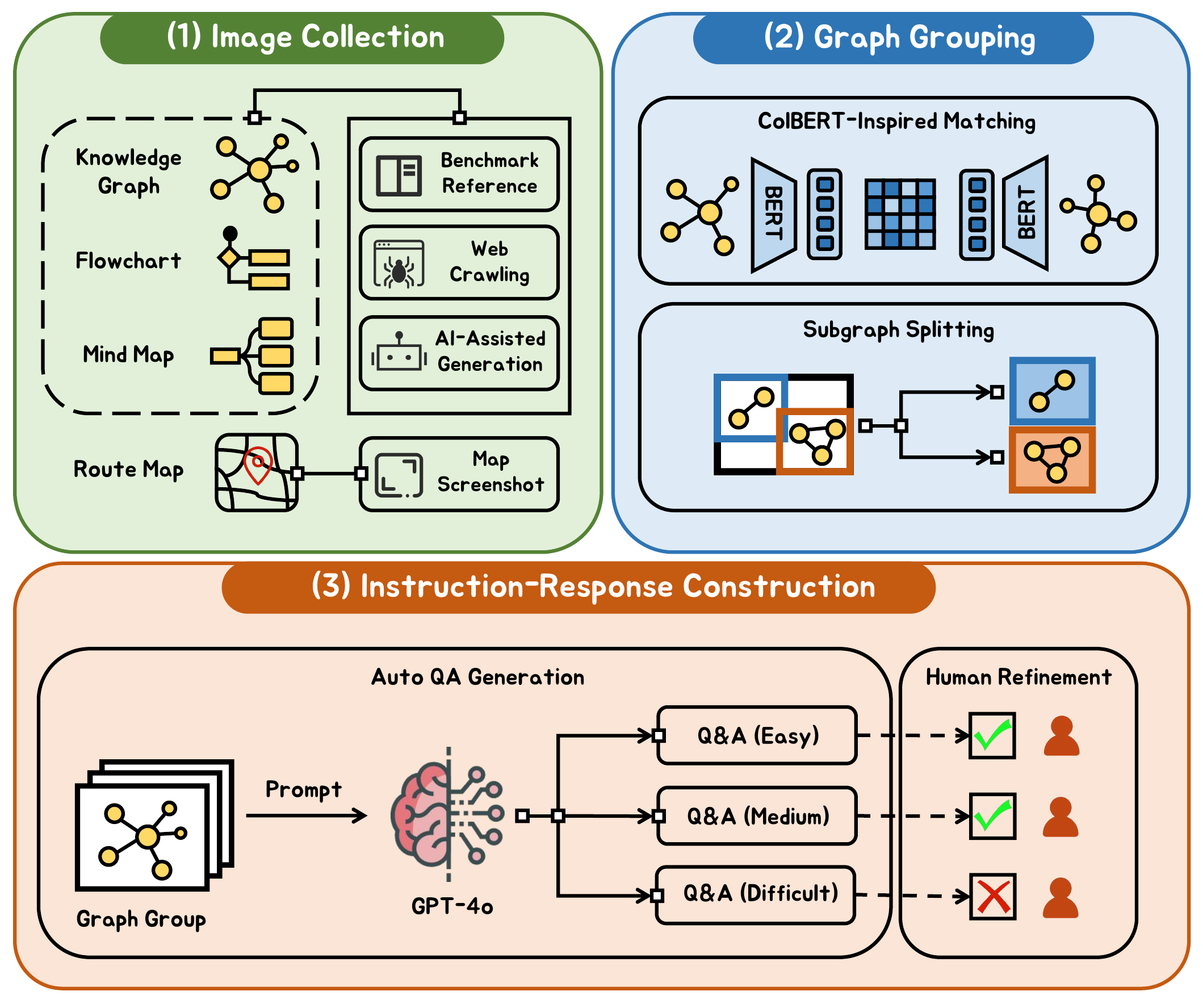}
    \caption{Overview of the benchmark construction pipeline. The process includes: (1) collecting diverse graph images across four types; (2) grouping them into semantically or structurally coherent sets using ColBERT-inspired matching or subgraph splitting; and (3) generating instruction-response pairs via GPT-4o, followed by manual review and refinement to ensure clarity and reasoning quality.}
    \label{figure3}
\end{figure}

The remainder of this section is organized as follows:
Section \ref{section3.2.0} details the image collection process, including the selection and preprocessing of graph images across different types.
Section~\ref{section3.2.1} describes the AI-assisted graph generation process used to enrich the domain-specific graph data.
Section \ref{section3.3} describes how these images are grouped into semantically or structurally related sets to support MGVR.
Section \ref{section3.4} presents our approach for generating instruction-response pairs using GPT-4o, tailored to promote cross-graph comprehension.
Finally, Section \ref{section3.5} presents a comprehensive statistical analysis of the benchmark, highlighting key characteristics and insights relevant to model evaluation.
The full data construction pipeline is illustrated in Figure \ref{figure3}.
\subsection{Graph Image Collection Strategies}
\label{section3.2.0}
We collected four types of graph images: knowledge graphs, flowcharts, mind maps, and route maps.
To ensure both diversity and quality, we employed a combination of data collection strategies, detailed as follows:

\textbf{Benchmark Referencing and Web Crawling.} 
We first obtained a large number of graph images by referencing the multimodal instruction-following benchmark \cite{ai2024advancement}, and further expanded the dataset by crawling web images using the benchmark’s associated keywords.
The keywords were generated by GPT-4o through extracting all nodes and edges from the benchmark graphs and summarizing them.
Benchmark referencing ensures high-quality and domain-relevant data, while web crawling enhances diversity by introducing a broader set of publicly available visual formats. 
These two methods are applicable to the first three graph types.

\textbf{AI-Assisted Graph Generation.} 
To further enrich knowledge graphs, flowcharts, and mind maps, we adopted an AI-assisted generation approach.
The full procedure, including prompt construction, keyword selection, and diagram generation, is detailed below.
\subsection{AI-Assisted Graph Generation}
\label{section3.2.1}
\begin{figure}[h]
    \centering
    \includegraphics[width=0.8\linewidth]{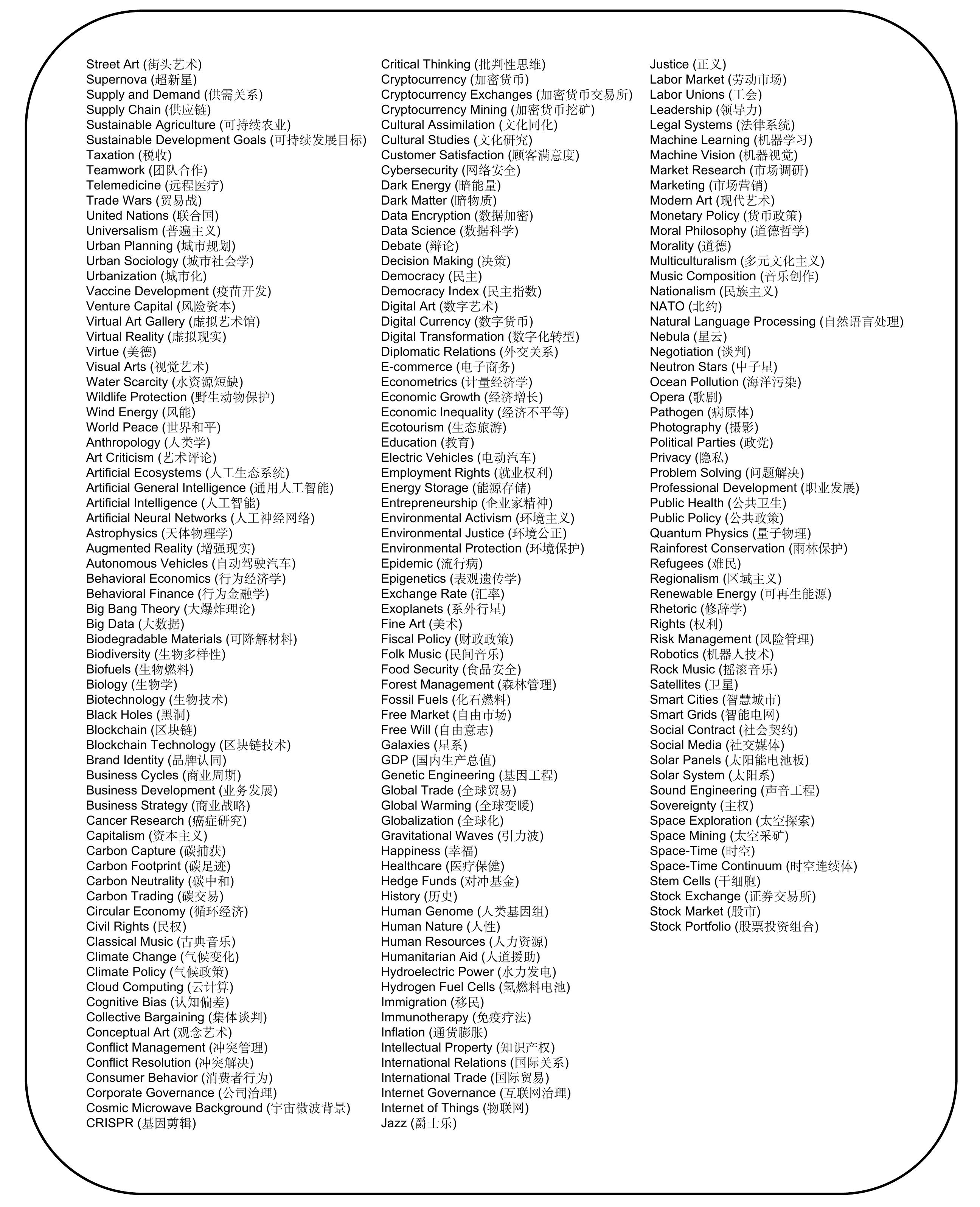}
    \caption{The 200 diverse domain-specific keywords generated by GPT-4o, which can be combined with the prompt in Figure \ref{keywords_prompt} to effectively guide GPT-4o in generating high-quality graph descriptions.}
    \label{keywords}
\end{figure}

\begin{figure}[h]
    \centering
    \includegraphics[width=0.9\linewidth]{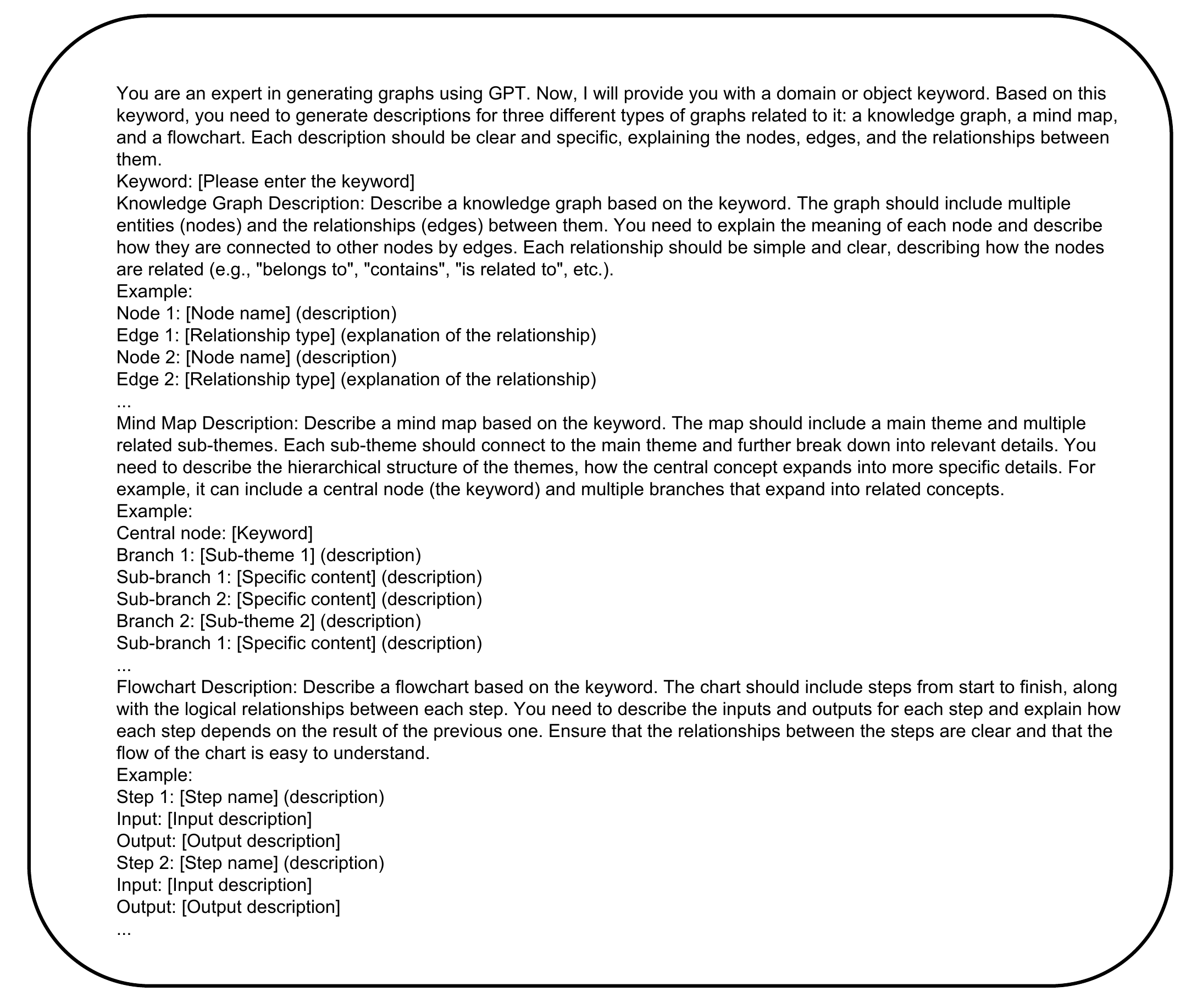}
    \caption{Prompt for guiding GPT-4o to separately generate descriptions of a knowledge graph, a mind map, and a flowchart based on a given keyword.}
    \label{keywords_prompt}
\end{figure}

\begin{figure}[h]
    \centering
    \includegraphics[width=\linewidth]{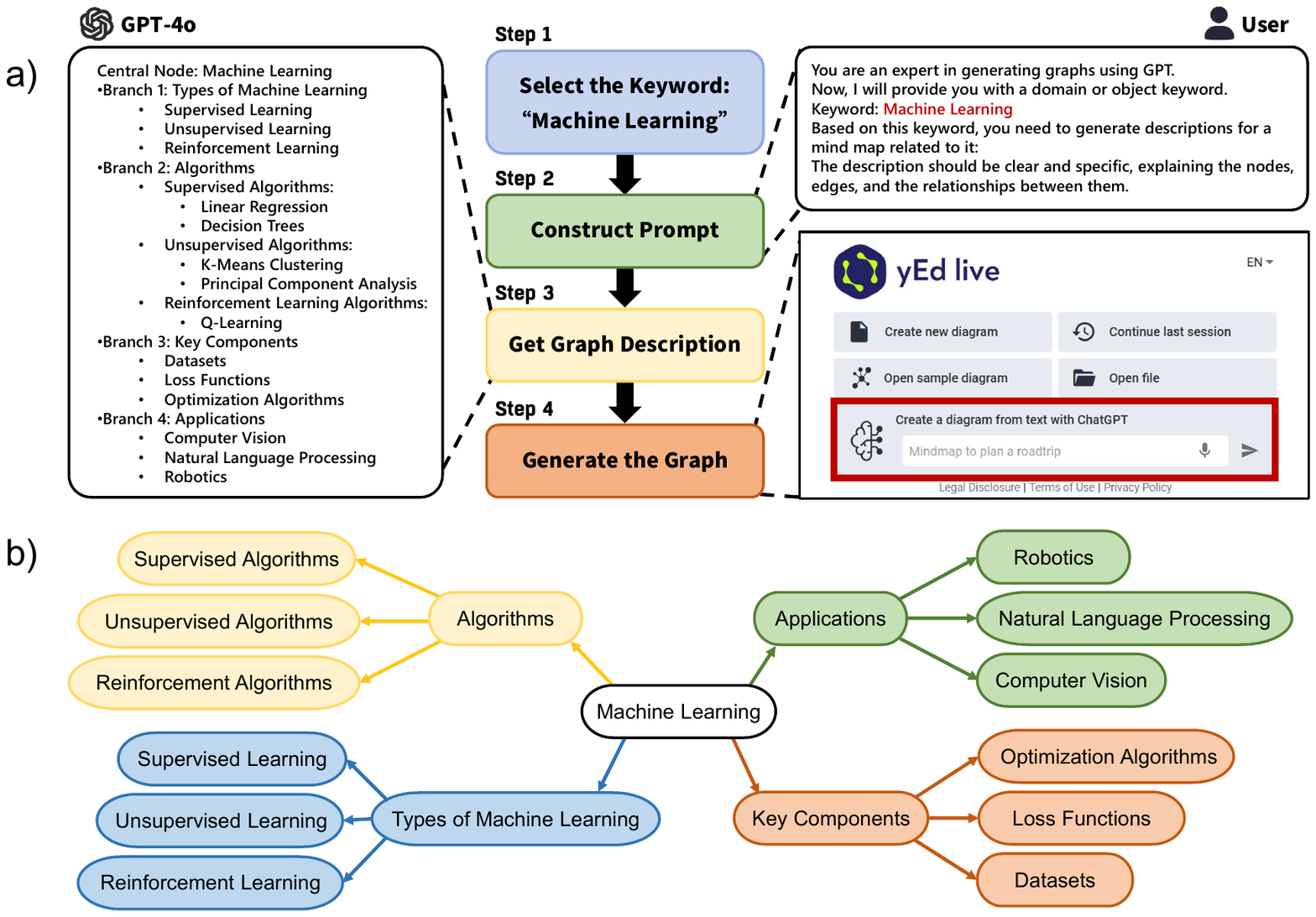}
    \caption{(a) The system pipeline for generating a mind map using the keyword “Machine Learning.” The process starts with keyword selection, followed by prompt construction, graph description generation, and automated mind map rendering using the yEd Live “Create a diagram from text with ChatGPT” tool. (b) An example of the generated mind map corresponding to the keyword.}
    \label{figure4}
\end{figure}

Specifically, we prompted GPT-4o to generate 200 diverse keywords spanning a wide range of domains. The resulting set of domain-specific keywords is illustrated in Figure \ref{keywords}.
For each keyword, GPT-4o was guided to produce a detailed graph description, including node names, edge labels, and connectivity information. 
The prompt used to guide this process is shown in Figure \ref{keywords_prompt}. 
These descriptions were input into the yEd Live drawing tool,\footnote{yEd Live is an online graph editor supporting automatic layout for structured graphs. (\url{https://www.yworks.com/yed-live/})} which then generated the corresponding graph images automatically.
For example, to generate a mind map on “Machine Learning”, we prompted GPT-4o to describe the graph, and used yEd Live’s “Create a diagram from text with ChatGPT” to produce the final diagram (see Figure \ref{figure4}).
Compared to directly selecting graphs from publicly available datasets, this method offers significant advantages:
Public datasets are often limited in scale, domain diversity, or relationship complexity. 
In contrast, AI-generated graphs can flexibly cover a wider range of topics and structures.
By leveraging carefully designed generation strategies, we achieved greater domain coverage, structural diversity, and complexity—enhancing the benchmark’s generality and its effectiveness in evaluating models’ cross-domain reasoning capabilities.

\textbf{High-Confidence Route Maps via Google Maps.} 
For route maps, we adopted a targeted strategy of capturing high-resolution screenshots from Google Maps.
This ensured the geographic accuracy and visual clarity of the maps, supporting more reliable downstream visual reasoning and inference.
\subsection{Grouping Graph Images by Semantic and Structural Relevance}
\label{section3.3}
To construct semantically coherent image groups, we employed tailored grouping strategies based on graph type.
Specifically, knowledge graphs, flowcharts, and mind maps—due to their conceptual overlap and structural compatibility—were grouped both within the same type and across different types.
In contrast, route maps, which primarily convey spatial and navigational information, differ fundamentally from the other categories. As such, they were grouped only within their own type to preserve thematic consistency and interpretability.

\subsubsection{ColBERT-Inspired Graph Grouping Strategy}
For the first three types of graphs, we adopted a \textbf{C}olBERT-\textbf{I}nspired \textbf{G}raph \textbf{G}rouping (\textbf{CIGG}) strategy, which leverages fine-grained token-level similarity to construct semantically meaningful image groups:

\begin{enumerate}

   \item \textbf{Graph Element Extraction:} We first designed prompts to leverage GPT-4o for extracting all node and edge names from each graph (see Appendix B of the supplementary material for prompt details). 
    
    \item \textbf{Semantic Encoding:} Each extracted node and edge name was encoded into semantic vectors using BERT, where the final-layer hidden states serve as the token-level representations.
    
    \item \textbf{ColBERT-Inspired Similarity Matching:} We adopt a bi-directional max-sim approach inspired by ColBERT~\cite{khattab2020colbert} to compute graph similarity, enabling the construction of semantically coherent graph groups. See Appendix B for details.    
\end{enumerate}
\subsubsection{Subgraph-Splitting Strategy for Route Maps}
Due to the high structural and semantic homogeneity among route maps, distinguishing them in the BERT semantic space is challenging.
Thus, the CIGG strategy is not suitable for this graph type.
To overcome this, we apply a subgraph-splitting strategy, detailed in Appendix B of the supplementary material.
\subsubsection{Manual Refinement}
Each constructed image group was manually reviewed, and those lacking meaningful semantic connections among the graphs were directly discarded. 
This step ensures that the remaining groups consist of graphs that are conceptually related and suitable for joint reasoning.
\subsection{Instruction-Response Construction}
\label{section3.4}
\subsubsection{VLM-based Instruction-Response Candidate Generation}
We carefully designed prompts to guide GPT-4o in generating effective instruction-response pairs for each group of graph images (see Appendix B of the supplementary material for prompt details).
For every graph group, GPT-4o is prompted to produce three instruction-response pairs with increasing difficulty levels: easy, medium, and difficult.
Each pair is required to involve reasoning across as many graphs in the group as possible, ensuring that the task truly reflects the challenge of MGVR.
\subsubsection{Manual Review and Refinement of Instruction-Response Pairs}

As a benchmark, the quality of samples is critical.
After initial generation, we conducted a rigorous manual review process to ensure each instruction-response pair met quality standards.
The full set of review criteria and editing actions are provided in Appendix B of the supplementary material.

\subsection{Data Statistics and Analysis}
\label{section3.5}
To assess the quality of our data, we randomly selected 10\% of the samples and invited independent annotators who were not involved in benchmark construction to evaluate the validity of the instruction-response pairs. All reviewed samples were deemed valid, further confirming the overall reliability of our benchmark.
In addition, we partitioned the dataset into training, validation, and test splits.
Importantly, the test set is carefully curated to ensure broad coverage of graph group types and difficulty levels, enabling a robust and balanced evaluation of model performance. An overview of the dataset is provided in Table~\ref{table1}. 
In total, the benchmark contains 4,021 instruction--response pairs, including 3,361 for training, 360 for validation, and 300 for testing. The validation split is balanced by construction: the four homogeneous graph group types each contribute exactly 20 clusters and the heterogeneous type contributes 40 clusters, and each cluster contributes one easy, one medium, and one difficult item, yielding a perfectly uniform difficulty distribution over its 360 items. The test split likewise contains exactly 100 easy, 100 medium, and 100 difficult items and is balanced across graph group types.
More benchmark statistics are provided in Appendix C.

\begin{table*}[!t]
  \caption{An overview of our MGVR benchmark.}
  \label{table1}
  \centering
  \begin{tabular}{ccccc}
\hline
& \textbf{\# Train} & \textbf{\# Valid} & \textbf{\# Test} & \textbf{\# Overall} \\ \hline
\textbf{Knowledge Graph-type} & 555 & 60  & 50  & 665  \\
\textbf{Flowchart-type}       & 555 & 60  & 50  & 665  \\
\textbf{Route Map-type}       & 552 & 60  & 50  & 662  \\
\textbf{Mind Map-type}        & 565 & 60  & 49  & 674  \\
\textbf{Heterogeneous-type}   & 1134 & 120 & 101 & 1355 \\
\textbf{Overall}             & 3361 & 360 & 300 & 4021 \\ \hline
\end{tabular}
\end{table*}

\section{Evaluation of Large Vision-Language Models on Multi-Graph Visual Reasoning}
\label{section4}
In this section, we present a comprehensive and systematic evaluation of several state-of-the-art VLMs on the proposed benchmark.

The evaluated models span both proprietary and open-source systems, including GPT-5.4 \cite{openai2026gpt54}(OpenAI), Gemini-3.1-Pro \cite{google2026gemini31pro} (Google DeepMind), Claude-Sonnet-4.6 \cite{anthropic2026claude}(Anthropic), Qwen3.5-122B-A10B\footnote{\url{https://huggingface.co/Qwen/Qwen3.5-122B-A10B}}, Qwen3.5-35B-A3B\footnote{\url{https://huggingface.co/Qwen/Qwen3.5-35B-A3B}} and Qwen3.5-27B\footnote{\url{https://huggingface.co/Qwen/Qwen3.5-27B}} \cite{qwen3.5}.

\subsection{What Abilities Do We Focus On?}
\label{section4.1}
Unlike conventional VQA, MGVR demands a richer evaluation protocol. We propose three dimensions to reflect its structural, semantic, and procedural complexity.

\textbf{Graph Parsing Accuracy (GPA).} 
This dimension evaluates the model’s ability to \emph{comprehend and interpret the structural features of graphs} and \emph{effectively apply this understanding within the context of the question}.
Accurate graph parsing is critical for successful MGVR.

\textbf{Reasoning Consistency and Completeness (RCC).} 
This dimension measures the logical consistency and completeness of the model’s reasoning process.
It reflects whether the model’s response demonstrates a \emph{coherent, well-structured, and internally consistent reasoning chain}.

\textbf{Instructional Reasoning Accuracy (IRA).} 
This dimension assesses whether the model can accurately follow the given instructions to \emph{generate correct or plausible answers}.
It directly reflects the model’s fundamental capacity for instruction-driven reasoning.

\subsection{Evaluation Strategy}
\label{section4.2}
In this section, we outline the evaluation strategy employed to assess model performance on the test set derived from our benchmark, as described in Section \ref{section3.5}. 
The test set consists of 300 samples, which encompass a diverse range of graph group types and reasoning difficulty levels.

We adopt a two-stage evaluation protocol. In the first stage, following the VLM-as-a-Judge paradigm, 
we employ Qwen3.8-27B\footnote{\url{https://huggingface.co/Qwen/Qwen3.8-27B}}, one of the most recent and capable open-source models, to assess model responses along the three dimensions defined in Section~\ref{section4.1}. Because three of the evaluated systems belong to the Qwen3.5 family, we further quantify the risk of same-family self-preference via a legacy GPT-4o cross-check in Appendix~E of the supplementary material. Specifically, we design a dedicated prompt for Qwen3.8-27B to evaluate each dimension on a five-point scale (1--5), where higher scores indicate better performance under the corresponding criterion. The full evaluation prompt is provided in Appendix E. For each response, the judge is provided with the original instruction, the complete set of graph images for the sample, and the reference answer.

In the second stage, we randomly select a subset of evaluation samples for human annotation. 
The human annotators are blinded to the automatic evaluation scores to eliminate bias. 
Importantly, the same evaluation dimensions and scoring criteria used in the automated evaluation are applied in the human assessment. 
Once the human evaluation is complete, we compute the correlation between the automatic evaluation scores and human judgments to assess the reliability and validity of the automated evaluation process.

\subsection{Evaluation Results and Analysis}
In this section, we present a comprehensive analysis of model performance from multiple perspectives. We summarize overall model performance in Section \ref{section4.4.1} and include human-machine consistency analysis in Section \ref{section4.4.2}. 
\subsubsection{Analysis of Overall Model Performance via Automatic Evaluation}
\label{section4.4.1}
\begin{table}[!ht]
  \caption{Overall model performance under automatic evaluation by Qwen3.8-27B on the test set. Bold indicates the best result.}
  \label{table2}
  \centering
  \begin{tabularx}{\columnwidth}{Xccc}
    \toprule
    &\textbf{IRA} & \textbf{GPA} & \textbf{RCC} \\
    \midrule
    \textbf{Qwen3.5-122B-A10B} & 4.06 & 4.02 & 4.08 \\
    \textbf{Qwen3.5-35B-A3B}   & 4.09 & 3.97 & 4.08 \\
    \textbf{Qwen3.5-27B}       & 4.15 & 4.06 & 4.15 \\
    \textbf{GPT-5.4}           & 4.21 & 3.84 & 4.11 \\
    \textbf{Gemini-3.1-Pro}    & 4.22 & \textbf{4.12} & \textbf{4.22} \\
    \textbf{Claude-Sonnet-4.6} & \textbf{4.28} & 4.03 & 4.20 \\
    \bottomrule
  \end{tabularx}
\end{table}

\begin{figure}[h]
    \centering
    \includegraphics[width=1.0\linewidth]{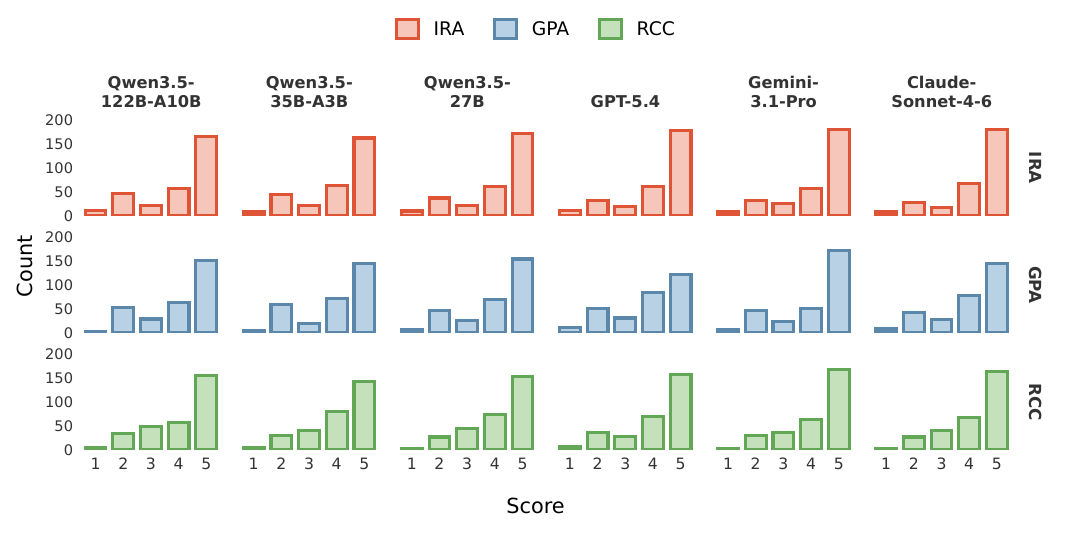}
    \caption{Overall model performance under automatic evaluation by the Qwen3.8-27B judge.}
    \label{figure8}
\end{figure}

Table \ref{table2} reports the average scores of each model across the three evaluation dimensions, while Figure \ref{figure8} shows the score distributions per dimension for each model.

We observe that Claude-Sonnet-4.6 achieves the highest IRA mean (4.28), whereas Gemini-3.1-Pro leads in GPA (4.12) and RCC (4.22). Among the three Qwen3.5 models, Qwen3.5-27B has the highest mean in all three dimensions (4.15, 4.06, and 4.15 for IRA, GPA, and RCC, respectively).
GPT-5.4 obtains an IRA mean of 4.21 but the lowest GPA mean (3.84), indicating a relative weakness in graph parsing under this evaluation. Across models, the means span 4.06--4.28 for IRA, 3.84--4.12 for GPA, and 4.08--4.22 for RCC.

\subsubsection{Human Evaluation and Consistency Analysis with Automatic Evaluation Scores}
\label{section4.4.2}

\begin{table}[!ht]
    \caption{Overall model performance under human evaluation, computed on the same 10\% human-labeled subsample as the consistency analysis in Section~\ref{section4.4.2}. Scores in parentheses denote the Qwen3.8-27B automatic scores on the same subsample. Bold indicates the best result.}
  \label{table3}
  \centering
  \begin{tabularx}{\columnwidth}{Xccc}
    \toprule
    &\textbf{IRA} & \textbf{GPA} & \textbf{RCC} \\
    \midrule
    \textbf{Qwen3.5-122B-A10B} & 4.32 (4.07) & 3.88 (4.00) & 4.44 (4.07) \\
    \textbf{Qwen3.5-35B-A3B} & 4.05 (4.10) & 3.84 (3.97) & 4.21 (4.10) \\
    \textbf{Qwen3.5-27B} & 4.26 (4.13) & 3.81 (4.03) & \textbf{4.57} (4.17) \\
    \textbf{GPT-5.4} & \textbf{4.38} (4.20) & 3.92 (3.87) & 4.56 (4.13) \\
    \textbf{Gemini-3.1-Pro} & 4.37 (4.23) & \textbf{3.93} (4.13) & 4.40 (4.20) \\
    \textbf{Claude-Sonnet-4.6} & 4.33 (4.30) & 3.86 (4.03) & 4.15 (4.23) \\
    \bottomrule
  \end{tabularx}
\end{table}

\begin{figure}[!ht]
    \centering
    \includegraphics[width=0.9\linewidth]{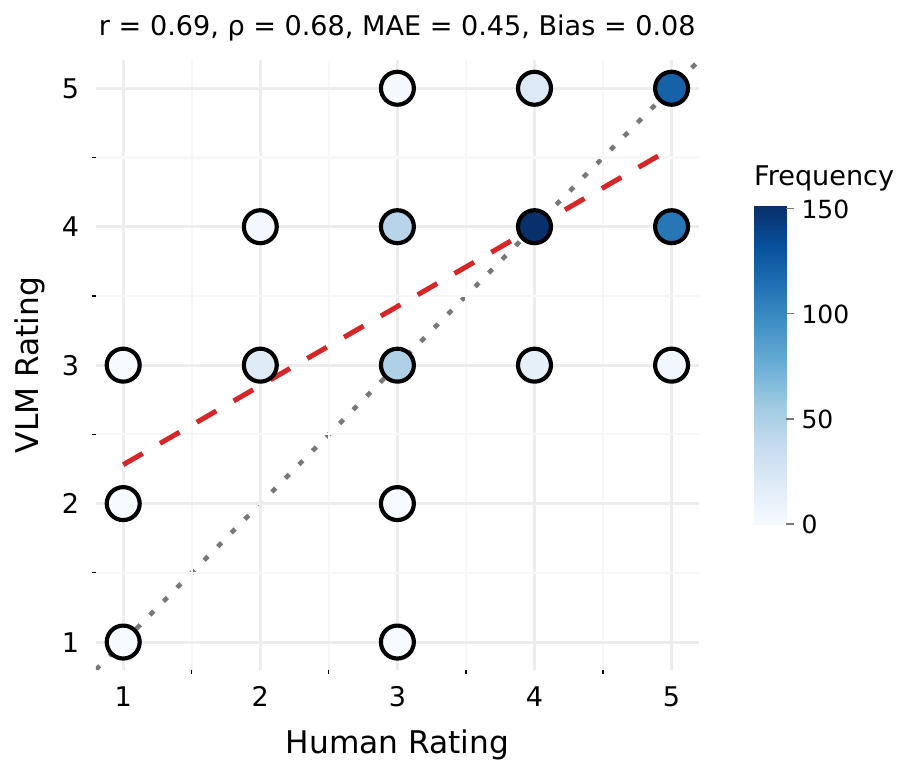}
    \caption{Scatter plot illustrating the consistency between automatic evaluation scores and human ratings across a 10\% randomly sampled subset of the evaluation dataset. 
    Each point represents the frequency of a specific score pair (Human, Qwen3.8-27B), with color intensity indicating occurrence frequency. The diagonal dotted line denotes perfect agreement, while the red dashed line shows the linear regression trend, representing the alignment between automatic evaluation scores and human scores—closer proximity to the diagonal indicates stronger consistency.}
    \label{figure11}
\end{figure}

To evaluate the consistency between automatic evaluation scores and human judgments, we randomly sampled 10\% of the evaluation dataset for manual annotation.
The results of the human evaluation are summarized in Table \ref{table3}, and a visualization of score consistency is provided in Figure \ref{figure11}.

We calculated four commonly used consistency metrics: the Pearson correlation coefficient (\(r\)), the Spearman rank correlation coefficient (\(\rho\)), the Mean Absolute Error (MAE), and Bias.
The results indicate a moderately strong linear correlation (\(r = 0.69\)) and a moderate rank correlation (\(\rho = 0.68\)) between automatic and human scores, suggesting that the automatic evaluator captures the overall scoring trends reasonably well, although some discrepancies remain in the ranking of individual samples. The mean absolute error (MAE) is 0.45, indicating that the average deviation from human scores is less than one point and remains within an acceptable range. The bias is \(+0.08\), showing that the automatic evaluator tends to slightly underestimate human scores.

To further characterize these discrepancies, we conduct a detailed dimension-wise analysis in Appendix~E of the supplementary material.

Despite these limitations, our carefully designed evaluation prompt enables the automatic scores to align well with human judgments in terms of overall trends.

\section{Ablation Studies}
\label{section:ablation}
The evaluation in Section \ref{section4} establishes how well current VLMs perform on MGVR, but not \emph{why}. In principle, a model could attain a high score without integrating information across graphs at all: by locating the answer in a single image, by transcribing the text printed in the figures, or by matching the surface form of the question to a template. This section isolates these possibilities. All ablation experiments use Qwen3.5-27B, the strongest open-source model in Table~\ref{table2}, as the subject model.

More detailed ablation results, including graph-type analyses showing that VLMs perform worst on route maps, are provided in Appendix E.

\subsection{Protocol}
\label{section:ablation_protocol}
In every condition, the only quantity that changes is the input given to the \emph{subject} model. The question text is byte-identical to the released test item, and the judge always receives the original question, the \emph{complete} original graph set, and the reference answer. A reply such as ``I can only see one diagram'' is therefore scored as a failure rather than credited as honesty, and no condition can raise its score by declining to answer. Decoding is greedy throughout.

Conditions that admit several realizations are enumerated exhaustively rather than sampled. For an item with $k$ graphs, the single-graph condition constructs all $k$ variants and averages them within the item before averaging across items; graph removal enumerates all $\binom{k}{m}$ position subsets; and graph substitution draws distractors of the same graph type from a different cluster using a per-item seed nested across $m$, so that consecutive levels share their noise instead of drawing fresh randomness. Because the test set contains only nine items with four or more graphs, the four- and six-image conditions are realized by padding the $k$ relevant graphs with irrelevant graphs of the same type, and are reported as ``$k$ relevant $+$ ($n-k$) irrelevant'' rather than as items that natively contain $n$ graphs.

To calibrate what counts as a change at all, we re-ran the subject model twice under identical settings. The two runs agree to within $+0.003$ GPA, $+0.007$ IRA and $-0.003$ RCC, with an item-level 95\% confidence half-width of $0.097$ across the 300 test items and $p\ge0.88$ in all three dimensions. We therefore treat $|\Delta|<0.10$ as the decoding-noise floor and do not interpret differences below it. All $p$-values in Table \ref{table:abl_main} come from paired Wilcoxon signed-rank tests with Holm correction within each dimension.

\begin{table*}[!t]
  \caption{Ablation of the subject model's input under the Qwen3.8-27B judge. $\Delta$ is the mean paired change from the reference input, with its 95\% confidence half-width; $d_z$ is the paired effect size; W/T/L counts GPA wins, ties and losses.}
  \label{table:abl_main}
  \centering
  \small
  \setlength{\tabcolsep}{4pt}
  \begin{tabular}{lccccccc}
\hline
\textbf{Subject model input} & \textbf{GPA} & $\boldsymbol{\Delta}$\textbf{GPA} & $\boldsymbol{p}$ & $\boldsymbol{d_z}$ & \textbf{W/T/L} & $\boldsymbol{\Delta}$\textbf{IRA} & $\boldsymbol{\Delta}$\textbf{RCC} \\ \hline
Full input ($k$ relevant graphs), reference & 4.080 & --- & --- & --- & --- & --- & --- \\
\hline
\multicolumn{8}{l}{\textit{Single-graph input}} \\
\quad best single graph of the $k$ & 3.120 & $-0.960\pm0.156$ & $<\!10^{-4}$ & $-0.70$ & 24/111/165 & $-1.080$ & $-0.663$ \\
\quad mean over the $k$ single-graph choices & 2.438 & $-1.642\pm0.139$ & $<\!10^{-4}$ & $-1.34$ & 11/26/263 & $-1.665$ & $-1.144$ \\
\quad first graph only & 2.607 & $-1.473\pm0.169$ & $<\!10^{-4}$ & $-0.99$ & 15/73/212 & $-1.587$ & $-1.070$ \\
\hline
\multicolumn{8}{l}{\textit{Graph removal (items with $k\ge3$)}} \\
\quad remove one graph, mean over choices & 3.341 & $-1.013\pm0.173$ & $<\!10^{-4}$ & $-1.03$ & 6/19/102 & $-0.823$ & $-0.630$ \\
\quad remove one graph, worst choice & 2.575 & $-1.780\pm0.231$ & $<\!10^{-4}$ & $-1.35$ & 1/21/105 & $-1.551$ & $-1.236$ \\
\hline
\multicolumn{8}{l}{\textit{Irrelevant graphs}} \\
\quad $k$ relevant $+$ irrelevant, 4 images, relevant first & 3.965 & $-0.105\pm0.107$ & 0.111 & $-0.11$ & 35/198/54 & $-0.139$ & $-0.059$ \\
\quad $k$ relevant $+$ irrelevant, 4 images, shuffled & 3.746 & $-0.324\pm0.125$ & $<\!10^{-4}$ & $-0.30$ & 29/175/83 & $-0.258$ & $-0.192$ \\
\quad $k$ relevant $+$ irrelevant, 6 images, relevant first & 3.819 & $-0.270\pm0.117$ & $<\!10^{-4}$ & $-0.26$ & 33/182/78 & $-0.222$ & $-0.164$ \\
\quad $k$ relevant $+$ irrelevant, 6 images, shuffled & 3.611 & $-0.478\pm0.142$ & $<\!10^{-4}$ & $-0.39$ & 34/150/109 & $-0.362$ & $-0.304$ \\
\quad all $k$ graphs replaced by irrelevant graphs & 1.287 & $-2.802\pm0.154$ & $<\!10^{-4}$ & $-2.10$ & 2/11/280 & $-2.495$ & $-1.942$ \\
\hline
\multicolumn{8}{l}{\textit{Text-mediated input}} \\
\quad structured text (nodes and edges), no image & 3.723 & $-0.357\pm0.136$ & $<\!10^{-4}$ & $-0.30$ & 36/171/93 & $-0.410$ & $-0.293$ \\
\quad transcription of all figure text, no image & 3.713 & $-0.367\pm0.131$ & $<\!10^{-4}$ & $-0.32$ & 34/173/93 & $-0.407$ & $-0.317$ \\
\quad images $+$ structured text & 4.070 & $-0.010\pm0.094$ & 0.879 & $-0.01$ & 43/207/50 & $-0.063$ & $-0.023$ \\
\quad no graph input & 1.513 & $-2.567\pm0.153$ & $<\!10^{-4}$ & $-1.90$ & 3/14/283 & $-2.137$ & $-1.820$ \\
\hline
\end{tabular}
\end{table*}

\subsection{Single-Graph, Graph-Removal and Irrelevant-Graph}
\label{section:ablation_shortcut}
Table \ref{table:abl_main} reports the full set of conditions, and Table \ref{table:abl_ladder} traces, level by level, the three input manipulations that the number of graphs makes possible.

Restricting the model to a single graph costs $1.64$ GPA, $1.66$ IRA and $1.14$ RCC across all 300 multi-graph items, and $1.90$, $1.74$ and $1.23$ on the 127 items with three or more graphs: the questions are not answerable from a single image. Even the most favorable choice of which graph to retain --- selecting, per item, the single graph that happens to score best --- still loses $0.96$ GPA, so no fixed single-image policy recovers the full-input score.

Withholding a single graph already costs $1.27$ GPA, $1.28$ IRA and $0.89$ RCC over all 300 items, and withholding two costs $1.83$, $1.69$ and $1.20$ on the 127 items with at least three graphs, so accuracy degrades as more of the graph set is hidden. On the nine items with four or more graphs, the full-input model attains a perfect $5.00/5.00/5.00$, and removing three of those graphs still costs $1.24$ GPA --- the answer cannot have come from a single image that happened to contain it. Removal is destructive at every depth, even though the model is free to answer from whatever remains, which is the behavior expected of genuine multi-graph reasoning rather than of single-image lookup.

Swapping one graph for an unrelated graph of the same type costs $1.24$ GPA, $1.02$ IRA and $0.80$ RCC on all 300 items, and $1.81$, $1.30$ and $1.11$ for two swaps on the 127 items with at least three graphs, with the number of input images held constant, so none of this is a context-length artifact. Replacement is about as damaging as outright removal at the same $m$ (remove one $-1.27$ vs.\ replace one $-1.24$ GPA; remove two $-1.83$ vs.\ replace two $-1.81$), which indicates that the model reads the \emph{specific} graphs the question refers to rather than pattern-matching on how many diagrams are present. Replacing all $k$ graphs is in fact worse than supplying no graph at all ($1.29$ vs.\ $1.51$ GPA), confirming that the score is driven by the content of the graph set and not by its presence. Taken together, these three manipulations close the concern from both sides: taking a required graph away hurts, and so does putting an equally sized irrelevant graph in its place.

\begin{table*}[!t]
  \caption{Effect of the number of graphs available to the subject model, under the Qwen3.8-27B judge. $\Delta$ is relative to the full input; ``Variants'' counts the distinct ablated inputs scored. The $\ge4$-graph rows cover only nine items.}
  \label{table:abl_ladder}
  \centering
  \small
  \begin{tabular}{lcccc}
\hline
\textbf{Setting} & \textbf{Variants} & $\boldsymbol{\Delta}$\textbf{GPA} & $\boldsymbol{\Delta}$\textbf{IRA} & $\boldsymbol{\Delta}$\textbf{RCC} \\ \hline
\multicolumn{5}{l}{\textit{Single graph shown}} \\
\quad items with $\ge2$ graphs & 743 & $-1.64$ & $-1.66$ & $-1.14$ \\
\quad items with $\ge3$ graphs & 397 & $-1.90$ & $-1.74$ & $-1.23$ \\
\quad items with $\ge4$ graphs & 43 & $-1.48$ & $-0.94$ & $-0.70$ \\ \hline
\multicolumn{5}{l}{\textit{Graphs removed}} \\
\quad remove 1, items with $\ge2$ graphs & 743 & $-1.27$ & $-1.28$ & $-0.89$ \\
\quad remove 2, items with $\ge3$ graphs & 439 & $-1.83$ & $-1.69$ & $-1.20$ \\
\quad remove 3, items with $\ge4$ graphs & 90 & $-1.24$ & $-0.88$ & $-0.66$ \\ \hline
\multicolumn{5}{l}{\textit{Graphs replaced by irrelevant graphs, image count fixed}} \\
\quad replace 1, items with $\ge2$ graphs & 743 & $-1.24$ & $-1.02$ & $-0.80$ \\
\quad replace 2, items with $\ge3$ graphs & 439 & $-1.81$ & $-1.30$ & $-1.11$ \\
\quad replace 3, items with $\ge4$ graphs & 90 & $-0.98$ & $-0.62$ & $-0.53$ \\ \hline
\end{tabular}
\end{table*}

\subsection{Number of Input Graphs}
\label{section:ablation_count}
\begin{figure*}[!t]
    \centering
    \includegraphics[width=\textwidth]{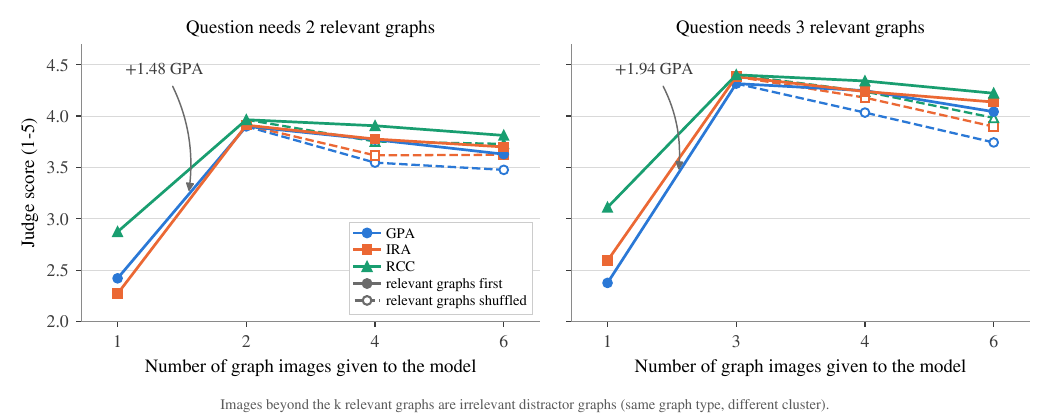}
    \caption{Score as a function of the number of graph images given to the subject model, under the Qwen3.8-27B judge. Left: items requiring two relevant graphs; right: three. Solid lines place the relevant graphs first, dashed lines shuffle them among the distractors.}
    \label{figure:imgcount}
\end{figure*}

Figure \ref{figure:imgcount} places these conditions on a common axis for items whose questions require two and three relevant graphs, respectively. The curve is dominated by a single step: moving from one graph to the $k$ graphs the question requires gains $1.48$ GPA at $k=2$ and $1.94$ GPA at $k=3$ ($p<10^{-4}$ in both cases). Beyond $k$, additional graphs are by construction irrelevant, and their cost depends more on where they are placed than on how many there are. Padding to four images with the relevant graphs kept first costs $0.11$ GPA and does not reach significance after correction ($p=0.111$), whereas interleaving the same distractors among the relevant graphs costs $0.32$ ($p<10^{-4}$); padding to six costs $0.27$ with the relevant graphs first, against $0.48$ when shuffled. The benchmark's difficulty therefore stems from the relevant graph set, and what the extra images probe is the model's ability to locate the graphs the question refers to --- a capability that degrades gracefully rather than collapsing.

\subsection{Text-Only: OCR and Structured Text}
\label{section:ablation_text}

Replacing graph images with a textual serialization reduces GPA by $0.36$ ($4.08 \rightarrow 3.72$). A verbatim transcription of all strings appearing in the figure---representing the upper bound of any OCR pipeline---achieves $3.71$, which is statistically indistinguishable from the explicit node--edge listing at $3.72$ ($p = 0.991$). This result indicates that the performance drop is not attributable to how the text is organized; rather, it reflects the absence of the visual modality itself. At the same time, both text-only baselines remain substantially above the $1.51$ GPA obtained without any graph input, confirming that these serializations preserve much of the task-relevant semantic content and that the observed gap is not an artifact of degenerate text input. Overall, the benchmark cannot be reduced to OCR followed by keyword matching over figure text.

Adding the same structured text to the images changes GPA by only $-0.01$ ($4.08 \rightarrow 4.07$, $p = 0.879$), which lies well within the noise floor, whereas adding images to the same text yields a gain of $+0.35$ GPA. This asymmetry indicates that the visual input largely subsumes the textual input: the information contributed by the serialization is already recoverable from the pixels, whereas the information contributed by the pixels---including edge labels, spatial routing, uncaptioned pictograms, and global figure layout---has no full counterpart in text. Taken together with the text-only results, this pattern is more consistent with native graph-structure parsing than with text-mediated retrieval: the model does not require an explicitly specified node--edge representation to reason over a diagram, and providing one offers no measurable benefit.

The Appendix E further analyzes examples that the subject model answers correctly from images alone and from images plus text, allowing the text-only conditions to isolate what each serialization fails to preserve.

\section{Fine-Tuning Open-Source Vision-Language Models}
\label{section5}
\textbf{Baseline Models.} 
We evaluate six representative lightweight models as our baselines: DeepSeek-VL-1.3B-Chat\footnote{\url{https://github.com/deepseek-ai/DeepSeek-VL}} \cite{lu2024deepseekvlrealworldvisionlanguageunderstanding}, InternVL2-1B, InternVL2.5-1B, InternVL2.5-1B-MPO\footnote{\url{https://github.com/OpenGVLab/InternVL}} \cite{chen2025expandingperformanceboundariesopensource}, Janus-1.3B\footnote{\url{https://github.com/deepseek-ai/Janus}} \cite{wu2024janusdecouplingvisualencoding}, and mPLUG-Owl3-1B-241014\footnote{\url{https://github.com/X-PLUG/mPLUG-Owl}} \cite{ye2024mplugowl3longimagesequenceunderstanding}.

\textbf{Experiment Details.}
We fine-tuned each model using the training set described in Section \ref{section3.5} and evaluated their performance on the same evaluation dataset introduced in Section \ref{section4.2}, enabling a direct comparison before and after fine-tuning.
The evaluation followed the protocol outlined in Section \ref{section4.1}, assessing model outputs along three dimensions: graph parsing accuracy (GPA), reasoning consistency and completeness (RCC), and instructional reasoning accuracy (IRA).
Each backbone was fine-tuned independently three times with identical hyperparameters, varying the random seed and data-shuffling seed across 42, 43, and 44. Each resulting adapter was evaluated on all 300 test items with greedy decoding.
Both the base model and its three fine-tuned variants were scored in the same evaluation round by the Qwen3.8-27B judge. This matched scoring setup supports within-judge before/after comparisons.
For each dimension, we report the mean and sample standard deviation (SD) of the three post-fine-tuning test-set means. We additionally report the mean paired item-level change ($\Delta$), its 95\% confidence interval over 300 items, and the paired Wilcoxon signed-rank $p$-value. These item-level intervals and tests are distinct from the across-seed SD; the three seed runs do not constitute 900 independent test items. The detailed results are provided in Appendix F of the supplementary material.

\subsection{Results}

\begin{table*}[!t]
  \caption{Scores before and after fine-tuning under the Qwen3.8-27B judge. After is mean $\pm$ SD across three training runs; bold marks the best After mean per dimension.}
  \label{table4}
  \centering
  \begin{tabular}{ccccccc}
\hline
& \multicolumn{2}{c}{\textbf{IRA}} & \multicolumn{2}{c}{\textbf{GPA}} & \multicolumn{2}{c}{\textbf{RCC}} \\ \cline{2-7} 
& \textbf{Before}          & \textbf{After}          & \textbf{Before}         & \textbf{After}         & \textbf{Before}         & \textbf{After}         \\ \hline
\textbf{DeepSeek-VL-1.3B-Chat} & 1.507 & $2.192\pm0.013$ & 1.243 & $1.704\pm0.036$ & 1.400 & $1.761\pm0.014$ \\
\textbf{InternVL2-1B}         & 2.127 & $2.509\pm0.010$ & 1.663 & $1.967\pm0.015$ & 1.793 & $2.010\pm0.010$ \\
\textbf{InternVL2.5-1B}       & 2.327 & $2.623\pm0.010$ & 1.853 & $\mathbf{2.131}\pm0.026$ & 2.030 & $2.161\pm0.021$ \\
\textbf{InternVL2.5-1B-MPO}   & 2.440 & $\mathbf{2.640}\pm0.030$ & 1.913 & $2.120\pm0.027$ & 2.167 & $\mathbf{2.167}\pm0.036$ \\
\textbf{Janus-1.3B}           & 2.003 & $2.228\pm0.012$ & 1.570 & $1.737\pm0.038$ & 1.670 & $1.821\pm0.047$ \\
\textbf{mPLUG-Owl3-1B-241014}  & 1.860 & $2.256\pm0.034$ & 1.443 & $1.722\pm0.005$ & 1.673 & $1.923\pm0.010$ \\ \hline
\end{tabular}
\end{table*}

\begin{figure}[h]
    \centering
    \includegraphics[width=1.0\linewidth]{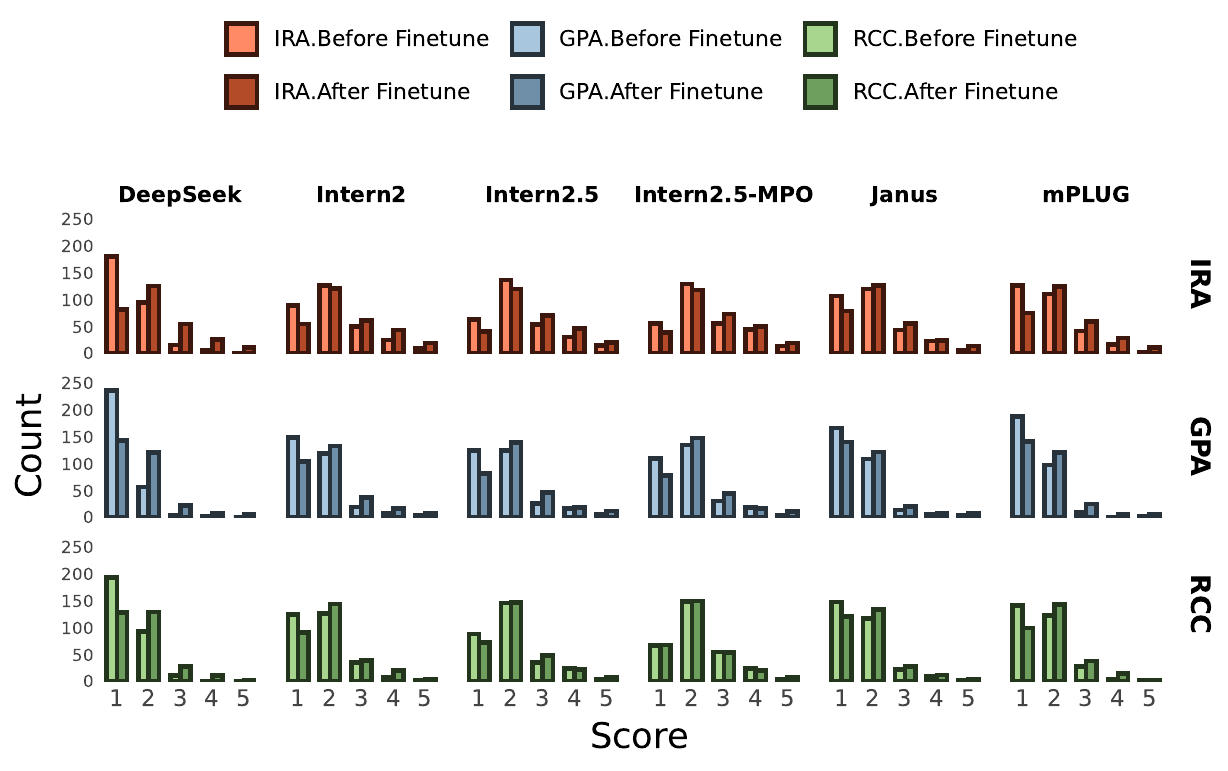}
    \caption{Score distributions for each VLM across three evaluation dimensions: GPA, RCC and IRA. Each row corresponds to a specific evaluation dimension, and each column to a different model. For each score level (1–5), lighter bars indicate results before finetuning, while darker bars represent results after finetuning.}
    \label{fig:ft_score_distribution}
\end{figure}

Table \ref{table4} reports before/after means, and Figure \ref{fig:ft_score_distribution} shows how the underlying score distribution shifts in each dimension. Fine-tuning produces positive mean gains in 17 of the 18 model--dimension combinations, all with reported paired Wilcoxon $p<0.05$; 13 have $p<10^{-4}$. These are individual-test $p$-values without a reported multiple-comparison correction. InternVL2.5-1B-MPO shows no detectable RCC gain ($\Delta=0.000$, 95\% CI $[-0.093,0.093]$, $p=0.762$), although its GPA and IRA improve.

DeepSeek-VL-1.3B-Chat achieves the largest mean gains in every dimension: $+0.461$ in GPA, $+0.686$ in IRA, and $+0.361$ in RCC (all $p<10^{-4}$). The largest across-seed SD among all reported post-fine-tuning means is 0.047, indicating limited variation in test-set averages across these three runs. This does not imply uniform gains for individual test items. In absolute post-fine-tuning scores, InternVL2.5-1B leads in GPA (2.131), while InternVL2.5-1B-MPO leads in IRA (2.640) and RCC (2.167).

Appendix F of the supplementary material reports all seed-level means, paired changes, confidence intervals, and $p$-values.

\section{Conclusion}
\label{section6}
This work investigates MGVR with VLMs as a paradigm that complements graph neural networks: instead of consuming an explicitly constructed node-and-edge representation, a VLM reads a diagram in its rendered form and reasons over the structure it perceives, thereby removing the need for an upstream graph-extraction stage. We accordingly scope this study to visual diagram question answering, where the task is to integrate structured information distributed across several rendered graphs.
We propose a benchmark that addresses gaps in both data and evaluation protocols.
Experiments show that although state-of-the-art VLMs exhibit strong potential, they still struggle with structural and semantic integration.
Fine-tuning lightweight open-source models on our benchmark improves 17 of the 18 model--dimension combinations we measure, confirming its effectiveness and generalizability as a training resource.

\section{Limitations}
Owing to the high computational cost of large VLMs, our fine-tuning is restricted to smaller models, which precludes a scalability analysis.
Moreover, the benchmark does not yet support non-visual formats such as tables, which limits its applicability to multimodal scenarios involving graph–table combinations.
Under the Qwen3.8-27B judge, Claude-Sonnet-4.6 leads in IRA, whereas Gemini-3.1-Pro leads in GPA and RCC. A sampled human evaluation confirms moderate agreement with the local judge (Section~\ref{section4.4.2}); the rankings remain judge-dependent and may shift under alternative protocols.
Fine-tuning gains are not uniform across dimensions: InternVL2.5-1B-MPO shows no detectable improvement in RCC despite gains in GPA and IRA. Nor does the low across-seed variation in average scores establish uniform gains for individual test items.
We do not report a graph neural network baseline. GNNs consume an explicit node-and-edge representation, which our setting deliberately withholds; providing one would require an oracle parse of every diagram and would therefore probe a different input assumption rather than the same task, so the two families are not directly comparable here.
Finally, the effects we measure are not uniform across graph group types. Route maps are consistently the lowest-scoring slices, whereas the sensitivity to withholding input graphs is largest for tasks that genuinely require several graphs and smallest for those that do not. Effect sizes measured on one slice should therefore not be extrapolated to another.

\FloatBarrier
\clearpage
\bibliographystyle{IEEEtran} 
\bibliography{custom}        
\newpage

\clearpage
\setcounter{figure}{0}
\setcounter{table}{0}
\setcounter{equation}{0}

\appendices

\section{Overview of Appendix}
\label{appendix:overview}
This supplementary material provides additional details, analyses, and ethical considerations that support the main paper. It is organized as follows:

\noindent \textbf{Additional details}
\begin{itemize}
    \item Appendix~\ref{appendix2}: Additional Benchmark Construction Details.
    \item Appendix~\ref{appendix:benchmark_stats}: Additional Benchmark Statistics.
    \item Appendix~\ref{appendix:implement_details}: Additional Implementation Details.
\end{itemize}

\noindent \textbf{Additional analysis}
\begin{itemize}
    \item Appendix~\ref{appdix:eval_analysis}: Additional Evaluation Analysis.
    \item Appendix~\ref{appendix:ft_analysis}: Additional Fine-Tuning Analysis.
\end{itemize}

\noindent \textbf{Ethical issue}
\begin{itemize}
    \item Appendix~\ref{appendix:ethics}: Ethical Considerations.
\end{itemize}

\section{Additional Benchmark Construction Details}
\label{appendix2}
\subsection*{GPT-4o Prompt for Graph Token Extraction}
\label{appendix2.1}
\begin{figure}[h]
    \centering
    \includegraphics[width=0.8\linewidth]{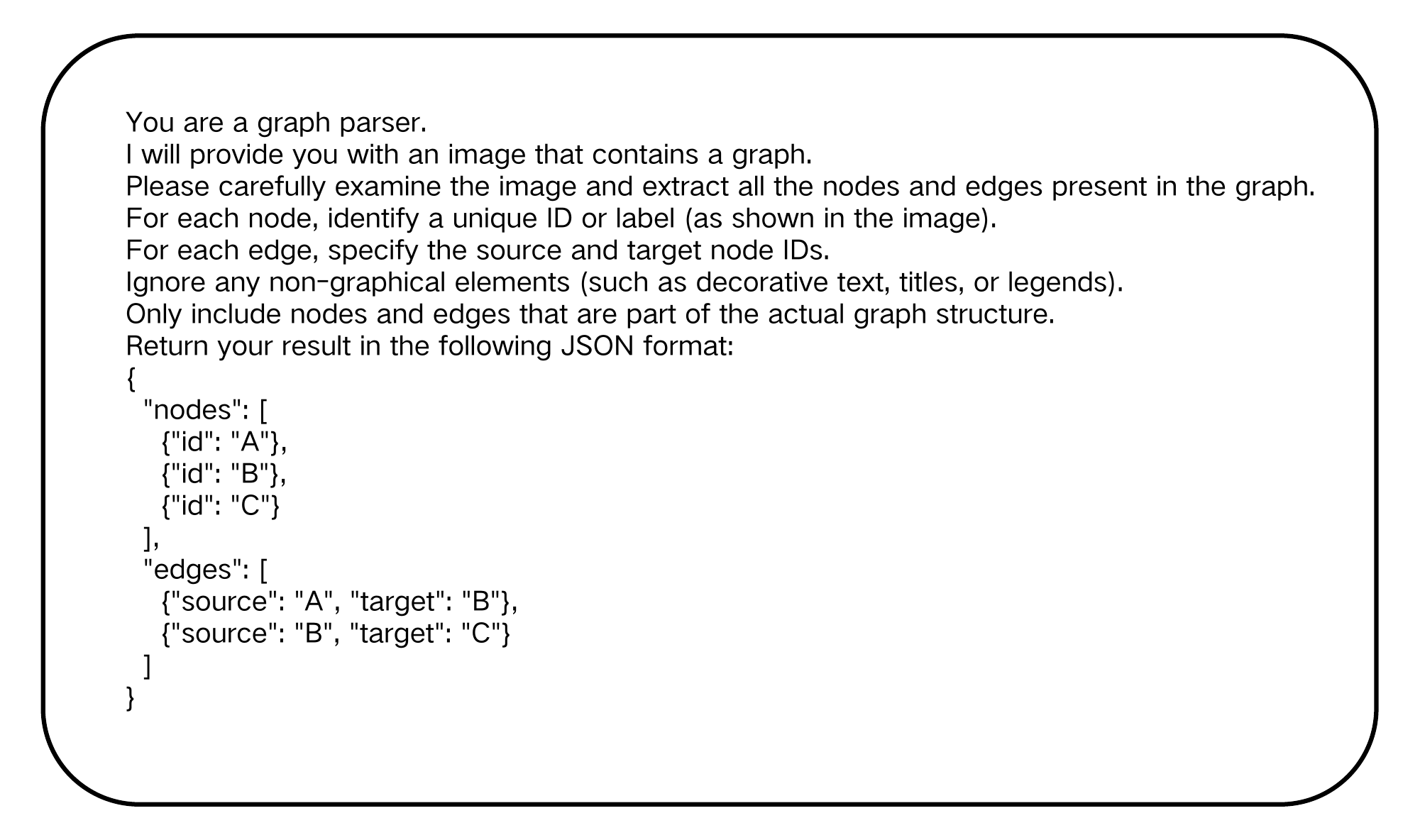}
    \caption{Prompt used to guide GPT-4o in parsing graph images into structured node and edge representations. The prompt instructs the model to identify all graphical nodes and edges from a given image while ignoring non-structural elements such as decorative text, titles, or legends.}
    \label{extract_graph_prompt}
\end{figure}
The prompt used to instruct GPT-4o to extract node and edge tokens from individual graph images is shown in Figure \ref{extract_graph_prompt}.
\subsection*{Bi-directional Max-Sim Similarity Computation}
\label{appendix2.2}
Let $G^A$ and $G^B$ be two graphs. From each graph, we extract the set of node and edge tokens:
\begin{equation}
    T^A = \{ t_1^A, t_2^A, \ldots, t_{n_A}^A \}, \quad T^B = \{ t_1^B, t_2^B, \ldots, t_{n_B}^B \}
\end{equation}
Each token $t_i$ is encoded into a dense vector $\mathbf{e}_i \in \mathbb{R}^d$ using a pretrained BERT encoder. Denote the resulting embeddings as:

\begin{equation}
    E^A = \{ \mathbf{e}_1^A, \ldots, \mathbf{e}_{n_A}^A \}, \quad E^B = \{ \mathbf{e}_1^B, \ldots, \mathbf{e}_{n_B}^B \}
\end{equation}

We define the \textit{max-sim} score from $A$ to $B$ as:
\begin{equation}
    \text{MaxSim}(A \rightarrow B) = \frac{1}{n_A} \sum_{i=1}^{n_A} \max_{j} \cos(\mathbf{e}_i^A, \mathbf{e}_j^B)
\end{equation}

Similarly, the reverse direction is computed as:
\begin{equation}
    \text{MaxSim}(B \rightarrow A) = \frac{1}{n_B} \sum_{j=1}^{n_B} \max_{i} \cos(\mathbf{e}_j^B, \mathbf{e}_i^A)
\end{equation}

The final \textit{bi-directional similarity score} is the average of both directions:
\begin{equation}
\begin{split}
    \text{BiMaxSim}(A, B) = & \frac{1}{2} [ \text{MaxSim}(A \rightarrow B) \\
    & + \text{MaxSim}(B \rightarrow A) ]
\end{split}
\end{equation}

We use $\text{BiMaxSim}(A, B)$ as the similarity metric to identify and group semantically aligned graph image sets. This approach captures asymmetric alignment and encourages mutual relevance between token sets from two graphs.
\subsection*{Subgraph-Splitting Strategy for Route Maps}
\label{appendix:split_strategy}

Due to the high structural and semantic homogeneity among different route maps, we empirically observed that distinguishing them within the BERT semantic space is challenging.
Consequently, the CIGG strategy is not suitable for this graph type, as it struggles to reliably identify route maps with semantically related content.

\begin{figure}[h]
    \centering
    \includegraphics[width=\linewidth]{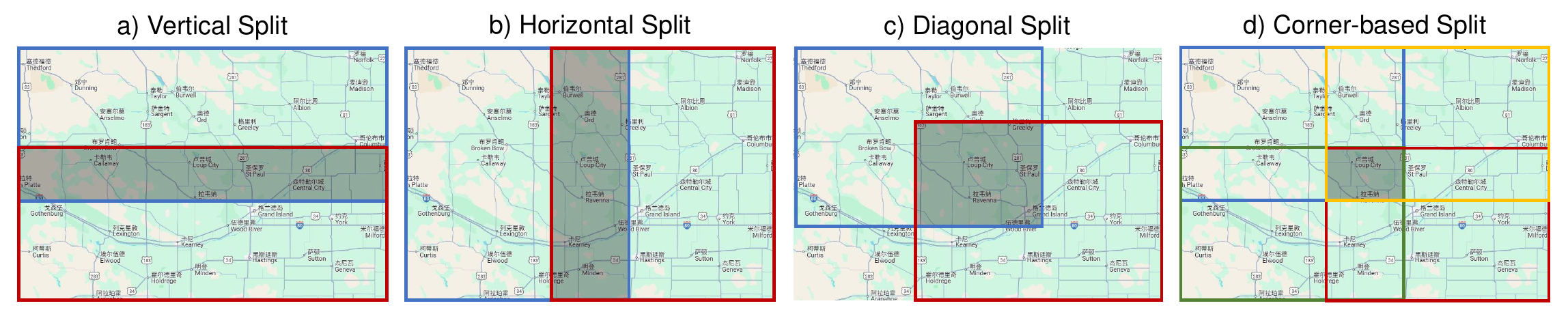}
    \caption{This figure illustrates four distinct methods for splitting route maps into subgraphs to generate image groups.
    \textbf{Vertical Split:} The map is split vertically into two subgraphs, marked by red and blue bounding boxes.
    \textbf{Horizontal Split:} The map is split horizontally into two subgraphs, indicated by red and blue boxes.
    \textbf{Diagonal Split:} The map is split diagonally into two overlapping subgraphs, shown in red and blue.
    \textbf{Corner-based Split:} The map is split into four subgraphs, each represented by yellow, green, blue, and red boxes.
    The shaded overlapping regions in each method indicate areas where subgraphs share common information, with the overlap controlled at 20\%–30\% of the original map’s area.}
    \label{figure5}
\end{figure}

To address this limitation, we adopt a subgraph-splitting strategy with two distinct configurations:

\begin{itemize}
    \item \textbf{Splitting into Two Subgraphs:} We adopt three splitting methods—vertical, horizontal, and diagonal—to divide a route map into two subgraphs (see Figure \ref{figure5} (a), (b), (c)). 
    \item \textbf{Splitting into Four Subgraphs:} We apply a corner-based splitting approach, segmenting the route map into four quadrants to generate four interrelated subgraphs (see Figure \ref{figure5} (d)).
\end{itemize}

These methods simulate different spatial partitions while preserving local structural coherence.

\subsection*{Prompt for Instruction-Response Pair Generation}
\label{appendix3}
\begin{figure}[h]
    \centering
    \includegraphics[width=0.9\linewidth]{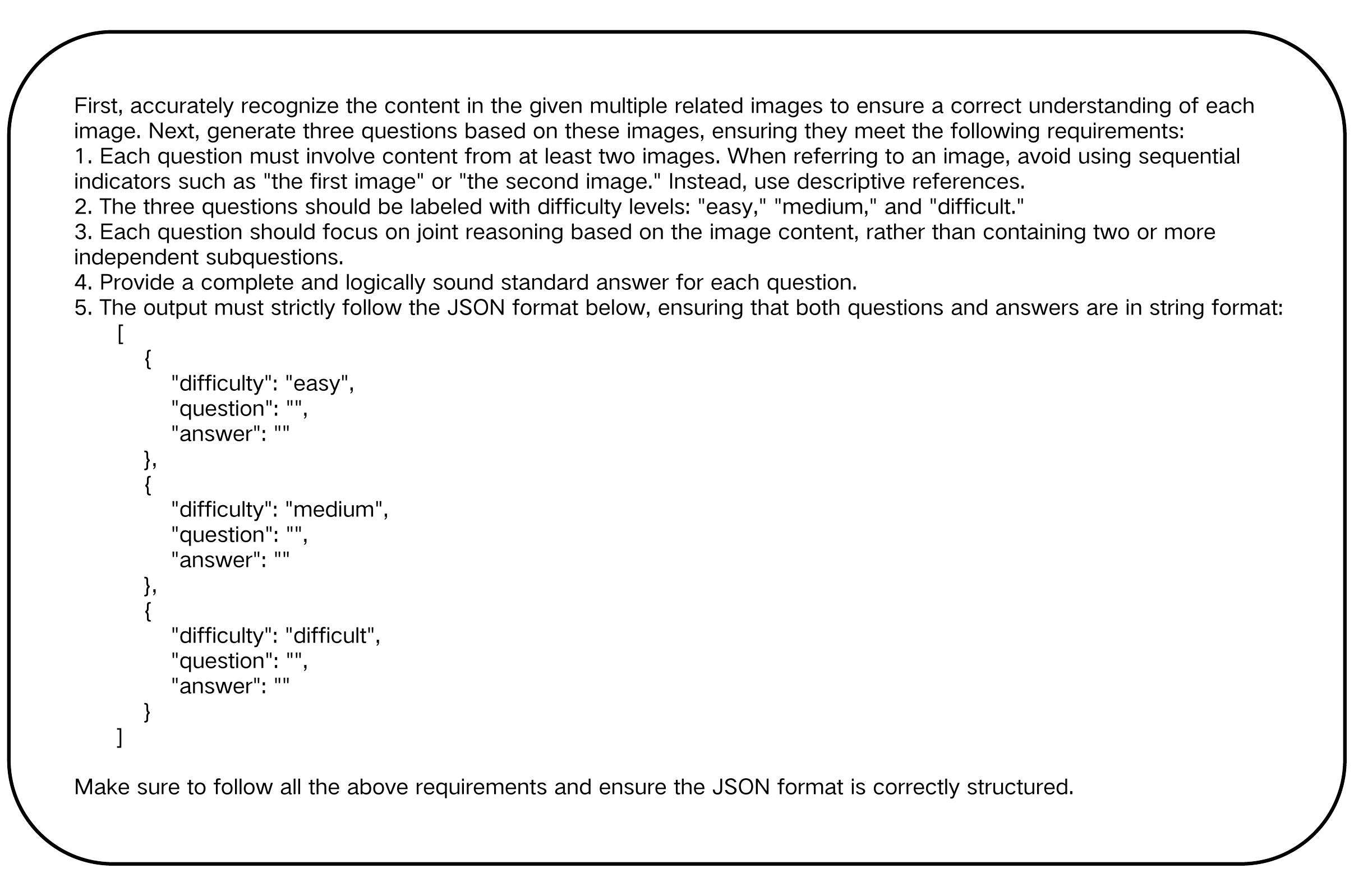}
    \caption{Prompt used to instruct GPT-4o in generating instruction-response pairs from multiple semantically related graph images.}
    \label{qa_prompt}
\end{figure}
The full prompt used to guide GPT-4o in generating instruction-response pairs for multi-graph visual reasoning (MGVR) tasks is shown in Figure \ref{qa_prompt}.

\subsection*{Manual Review and Refinement Procedure}
\label{appendix:manual_review}
To ensure the quality and consistency of instruction-response pairs, we performed a detailed manual review following the initial generation by GPT-4o.
This review process aimed to filter out low-quality samples and correct flawed responses to ensure the benchmark faithfully assesses multi-graph reasoning capabilities.

We identified three main reasons for deeming a sample invalid:
\begin{itemize}
    \item The instruction only referenced a single graph, violating the benchmark’s goal of assessing joint reasoning across multiple graphs.
    \item The instruction was irrelevant or unrelated to the graph content, resulting in semantically meaningless samples.
    \item The response contained factual inaccuracies or logical inconsistencies despite the instruction being valid.
\end{itemize}

Based on the above criteria, we applied the following corrective actions:
\begin{itemize}
    \item Pairs with invalid instructions were entirely removed from the benchmark.
    \item Pairs with valid instructions but flawed responses were manually edited to correct errors while preserving the original reasoning intent.
\end{itemize}

This annotation process  contributed significantly to the reliability of the released benchmark.

\section{Additional Benchmark Statistics}
\label{appendix:benchmark_stats}

\subsection*{Split Construction and Disjointness Verification}
\label{appendix:split_disjointness}

The benchmark is partitioned at the level of graph clusters. A cluster is the atomic unit of the dataset, consisting of a set of graph images together with all associated instruction--response pairs. Assigning entire clusters to splits ensures that no image group, and thus no individual image, is shared across splits, and that multiple instructions written for the same group are never distributed between the training and test portions of the benchmark.

The test set was constructed first and contains 300 QA pairs from 277 clusters, sampled to ensure balanced coverage across six graph types and three difficulty levels. After the test clusters were selected, they were excluded entirely from both the data-augmentation sampling pool and the train/validation split pool. The remaining clusters were then divided into training and validation sets, yielding 3,361 QA pairs from 1,121 clusters for training and 360 QA pairs from 120 clusters for validation. The validation set is strictly balanced, with 20 clusters for each of the four homogeneous types and 40 clusters for the heterogeneous type, and one QA pair per difficulty level within each cluster, thereby ensuring uniform coverage over all type--difficulty combinations.

We further performed programmatic verification of split disjointness. First, there is zero cluster overlap between any pair of splits. Second, there is zero duplicated question text among all training and validation QA pairs and between these QA pairs and the test set. Third, for the topic-organized subset (heterogeneous compositions), each topic directory is treated as an atomic cluster; therefore, topic-level disjointness follows directly from cluster-level disjointness. Fourth, the 300 newly augmented clusters are always assigned to the same split as their source clusters, which in practice is the training split, and therefore introduce no cross-split image reuse.

Table~\ref{table:split_disjointness} summarizes the resulting split statistics. All three pairwise cluster intersections are empty, as are the three pairwise intersections of the referenced image files. The 3,721 instruction instances in the training and validation splits contain no duplicate question text, and none overlaps with any of the 300 test questions. In addition, clusters sharing the same derivation are always assigned to the same split, ensuring that no training cluster is a variant of any test cluster.

\begin{table}[h]
  \caption{Split composition and disjointness. All pairwise intersections between splits are empty.}
  \label{table:split_disjointness}
  \centering
  \begin{tabular}{lccc}
\hline
& \textbf{Train} & \textbf{Valid} & \textbf{Test} \\ \hline
Instruction--response pairs & 3361 & 360 & 300 \\
Graph clusters             & 1121 & 120 & 277 \\
Graph image files          & 2748 & 294 & 685 \\ \hline
\multicolumn{4}{l}{\textit{Pairwise intersections}} \\
Clusters (train/valid, train/test, valid/test) & 0 & 0 & 0 \\
Image files (train/valid, train/test, valid/test) & 0 & 0 & 0 \\ \hline
\end{tabular}
\end{table}

The test split is further balanced with respect to graph-group type and difficulty level. The validation split is used only for checkpoint selection during fine-tuning and is never evaluated in any of the experiments reported in this paper.

\section{Additional Implementation Details}
\label{appendix:implement_details}
To support reproducibility and scalability analysis, we provide details of the compute environment for all experiments conducted in this work.

\textbf{Compute Environment:}
\begin{itemize}
    \item \textbf{GPU:} 4 $\times$ NVIDIA A100-SXM-80GB
    \item \textbf{CPU:} 62 vCPUs
    \item \textbf{RAM:} 990 GB
    \item \textbf{CUDA:} 12.4
    \item \textbf{OS/Container:} Ubuntu 22.04
    \item \textbf{Python Version:} 3.12
    \item \textbf{Frameworks:} PyTorch 2.5.1, ms-swift, vLLM
\end{itemize}

Table~\ref{tab:ft_hparams} lists the LoRA fine-tuning hyperparameters we apply, unchanged, to every one of the six baselines discussed in Section~V of the main paper. All values are read verbatim from a single ms-swift SFT driver script, invoked once per (backbone, seed) triple, so the ``identical hyperparameters'' clause of Section~V of the main paper refers to every field in the table simultaneously; deviations from the upstream ms-swift LoRA-SFT reference are explicitly marked.

\begin{table}[!t]
  \caption{LoRA fine-tuning hyperparameters shared by every baseline reported in Section~V of the main paper. Values marked \emph{(ours)} deviate from the ms-swift LoRA-SFT reference script; all other flags are ms-swift 3.12.6 defaults inherited verbatim.}
  \label{tab:ft_hparams}
  \centering
  \small
  \setlength{\tabcolsep}{4pt}
  \begin{tabular}{ll}
  \hline
  \textbf{Setting} & \textbf{Value} \\ \hline
  Framework & ms-swift 3.12.6 (PyTorch 2.5.1) \\
  Tuner type & LoRA \\
  LoRA rank $r$ & 8 \\
  LoRA alpha $\alpha$ & 32 \\
  Target modules & \texttt{all-linear} \\
  Learning rate & $2\!\times\!10^{-4}$ \emph{(ours; default $1\!\times\!10^{-4}$)} \\
  LR warmup ratio & 0.05 \\
  Precision & bfloat16 (\texttt{torch\_dtype}) \\
  Per-device train batch size & 1 \\
  Per-device eval batch size & 1 \\
  Gradient accumulation steps & 16 \\
  Effective batch size & 16 \\
  Max sequence length & 16{,}384 \emph{(ours; default 2{,}048)} \\
  Num.\ training epochs & 1\textsuperscript{$\dagger$} \\
  Random seeds & 42, 43, 44 (\texttt{seed}=\texttt{data\_seed}) \\
  Hardware (per fine-tuning run) & 1 $\times$ A100-SXM-80GB \\
  \hline
  \multicolumn{2}{p{0.95\linewidth}}{\footnotesize $\dagger$ mPLUG-Owl3-1B-241014 additionally receives a 3-epoch variant; it is the only backbone whose best evaluation loss lands on the final optimiser step of the 1-epoch schedule, indicating that its loss curve had not yet plateaued at the default budget. The 3-epoch run is reported alongside its 1-epoch counterpart in Appendix~\ref{appendix:ft_analysis}. The five remaining baselines are trained for 1 epoch only.} \\
  \end{tabular}
\end{table}

The ms-swift default \texttt{max\_length} of 2{,}048 tokens is insufficient once InternVL-family templates tile 2--6 graph images at up to 12 tiles $\times$ 256 tokens per tile, which under the default \texttt{truncation\_strategy=`delete'} silently drops the majority of training rows. Raising \texttt{max\_length} to 16{,}384 covers the top of the measured token-length distribution on our training and validation splits, and is applied uniformly across all six backbones for comparability. Consistent with the description in Section~V of the main paper, the two data-affecting seeds (\texttt{--seed} and \texttt{--data\_seed}, which ms-swift treats independently) are always set to the same value, so that the three repeated runs differ jointly in data-shuffling order and in LoRA initialisation.

\section{Additional Evaluation Analysis}
\label{appdix:eval_analysis}

To provide a more comprehensive analysis of both human and automatic evaluation, we present a dimension-wise bias analysis comparing human judgments with automatic scores, together with a detailed ablation study of factors that may affect evaluation.

Specifically, we examine the effects of multiple factors, including the number of input images, OCR-based cues, irrelevant graphs, and different graph reasoning task types. All ablation experiments are conducted using Qwen3.5-27B, which achieved the strongest performance among the open-source models in the main evaluation.

\subsection*{Same-Family Judge Risk}
Because Qwen3.8-27B serves as the automatic judge in Section~IV.B of the main paper while three of the six evaluated systems belong to the Qwen3.5 family, self-preference along model lineage is a legitimate concern. As a legacy pre-revision cross-check on the identical 300 test outputs, we compare our Qwen3.8-27B ratings against ratings from a GPT-4o judge (\texttt{gpt-4o-2024-08-06}). Averaged over IRA, GPA, and RCC, Qwen3.8-27B assigns \emph{lower} means than GPT-4o for every Qwen3.5 subject (gaps of $-0.15$, $-0.17$, and $-0.21$ for Qwen3.5-27B, Qwen3.5-35B-A3B, and Qwen3.5-122B-A10B), whereas the same gap on non-Qwen subjects is markedly smaller ($-0.04$ for Gemini-3.1-Pro, $-0.05$ for Claude-Sonnet-4.6, $-0.10$ for GPT-5.4). If same-family favoritism were present, the direction would be reversed. We therefore retain Qwen3.8-27B as the main-paper judge (Section~IV.B of the main paper): it is a purpose-built multimodal model that ingests the full graph set together with the reference answer, and it is substantially stronger than the earlier GPT-4o release on graph-heavy visual reasoning benchmarks such as MathVision and CharXiv.

\subsection*{Prompt for Automatic Evaluation}

\begin{figure}[!ht]
    \centering
    \includegraphics[width=\linewidth]{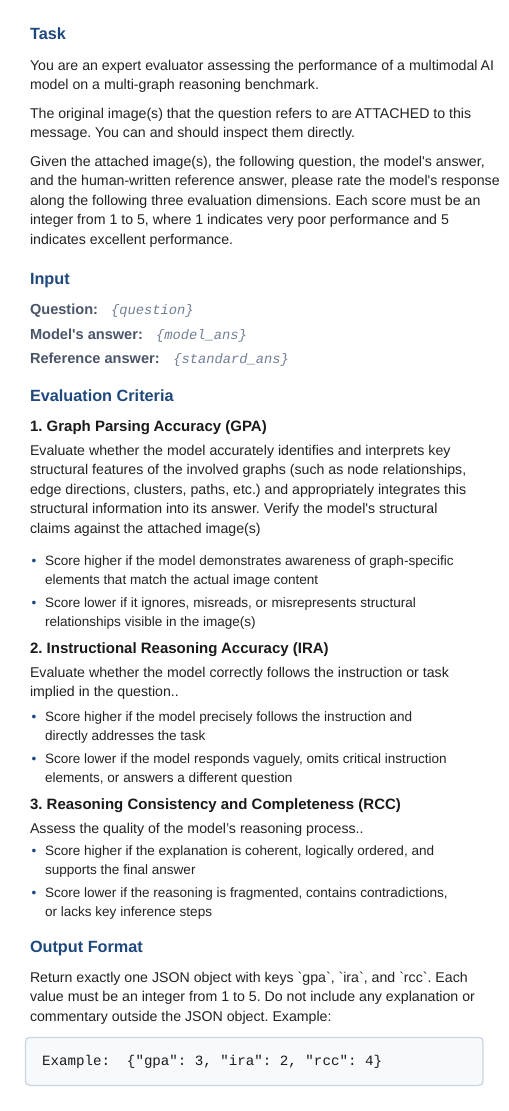}
    \caption{Prompt used to instruct Qwen3.8-27B to evaluate model-generated answers based on three core reasoning criteria: Graph Parsing Accuracy (GPA), Instructional Reasoning Accuracy (IRA), and Reasoning Consistency and Completeness (RCC). The prompt defines each criterion with specific expectations and provides scoring guidance to ensure consistent and fine-grained evaluation.}
    \label{evaluation_prompt}
\end{figure}

The prompt used to instruct Qwen3.8-27B to evaluate the model performance is provided in Figure~\ref{evaluation_prompt}.

\subsection*{Formal Definitions of Consistency Metrics}
\label{appendix7}
To quantify the consistency between automatic model scoring and human annotations, we adopt four commonly used statistical metrics: the \textbf{Pearson correlation coefficient} ($r$), the \textbf{Spearman rank correlation coefficient} ($\rho$), the \textbf{Mean Absolute Error} (MAE), and \textbf{Bias}. Their formal definitions and computation formulas are provided below. Throughout these definitions, we denote $x_i$ as the human-assigned score and $y_i$ as the corresponding model-assigned score.
\subsubsection{Pearson Correlation Coefficient ($r$)}
The Pearson correlation measures the linear relationship between human and model scores. Given paired scores $\{(x_i, y_i)\}_{i=1}^n$, it is defined as:
\begin{equation}
    r = \frac{\sum_{i=1}^n (x_i - \bar{x})(y_i - \bar{y})}{\sqrt{\sum_{i=1}^n (x_i - \bar{x})^2} \cdot \sqrt{\sum_{i=1}^n (y_i - \bar{y})^2}}
\end{equation}
where $\bar{x}$ and $\bar{y}$ denote the means of $x_i$ (human) and $y_i$ (model), respectively.

\subsubsection{Spearman Rank Correlation Coefficient ($\rho$)} 
Spearman's $\rho$ measures the monotonic relationship between the rankings of human and model scores. It is computed as:
\begin{equation}
    \rho = r(\text{rank}(x), \text{rank}(y))
\end{equation}
where $\text{rank}(x_i)$ and $\text{rank}(y_i)$ denote the ranks of the human and model scores, respectively.

\subsubsection{Mean Absolute Error (MAE)}  
MAE evaluates the average magnitude of the absolute differences between model and human scores:
\begin{equation}
    \text{MAE} = \frac{1}{n} \sum_{i=1}^n |y_i - x_i|
\end{equation}

\subsubsection{Bias}
Bias captures the average signed deviation of human scores from model scores, so that a positive value indicates that the automatic evaluator underestimates human ratings:
\begin{equation}
    \text{Bias} = \frac{1}{n} \sum_{i=1}^n (x_i - y_i)
\end{equation}

These metrics together provide a comprehensive view of the alignment between automatic evaluation and human judgment.

\subsection*{Dimension-Wise Bias Analysis between Human and Automatic Evaluation}
\label{appendix:dim_bias_analysis}

A closer examination of the dimension-wise bias reveals a clear asymmetric pattern: human ratings are higher than automatic scores for IRA and RCC, whereas the automatic judge assigns higher scores than human annotators for GPA. This discrepancy likely reflects differences in evaluation emphasis between human and automatic assessment.

The largest divergence is observed in RCC (avg.\ bias $= +0.24$), where human annotators consistently rate reasoning consistency and completeness more favorably than the Qwen3.8-27B judge. One possible explanation is that the automatic evaluator penalizes responses whose reasoning, although internally coherent, deviates from the logical structure of the reference answer or omits intermediate steps explicitly rewarded by the judging prompt. Human annotators, by contrast, may place greater weight on the overall coherence, plausibility, and communicative effectiveness of the reasoning, leading to more favorable judgments.

By contrast, GPA is the only dimension in which automatic scores exceed human ratings (avg.\ bias $= -0.13$). This pattern suggests that the Qwen3.8-27B judge may be relatively lenient toward surface-level structural correctness in graph parsing, whereas human annotators are more likely to penalize responses that identify graph elements correctly but fail to demonstrate a deeper understanding of their semantic relations. IRA shows a modest positive bias ($+0.11$), indicating that human evaluators assign slightly higher scores for instruction-following accuracy, possibly because they value practical usefulness and communicative adequacy beyond strict conformity to the reference answer.

Overall, these findings highlight the complementary roles of human and automatic evaluation in assessing VLM performance.

\subsection*{Evaluation Across Number of Input Graphs}
\label{appendix:imgcount}

The main paper plots GPA against the number of graph images supplied to the subject model; Table~\ref{table:appC_imgcount} lists the underlying values for all three dimensions. Within each block the item set is held fixed and only the images placed in the subject model's prompt vary; the judge always receives the complete original graph set, the original question and the reference answer, so that a response that ignores or misreads part of the evidence is still scored against the full evidence.

\begin{table}[h]
  \caption{Scores under the Qwen3.8-27B judge against the number of graph images given to the subject model. $^{\dagger}$The two orderings coincide, so these rows appear once.}
  \label{table:appC_imgcount}
  \centering
  \small
  \setlength{\tabcolsep}{4pt}
  \begin{tabular}{lcccccc}
\hline
 & \multicolumn{3}{c}{\textbf{Relevant first}} & \multicolumn{3}{c}{\textbf{Shuffled}} \\ \cline{2-4} \cline{5-7}
\textbf{\# images} & \textbf{GPA} & \textbf{IRA} & \textbf{RCC} & \textbf{GPA} & \textbf{IRA} & \textbf{RCC} \\
\hline
\multicolumn{7}{l}{\textit{Questions requiring $k=2$ relevant graphs ($n=170$)}} \\
1$^{\dagger}$ & 2.42 & 2.27 & 2.87 & --- & --- & --- \\
2$^{\dagger}$ (all relevant) & 3.90 & 3.91 & 3.96 & --- & --- & --- \\
4 & 3.77 & 3.78 & 3.91 & 3.55 & 3.62 & 3.75 \\
6 & 3.63 & 3.70 & 3.81 & 3.48 & 3.62 & 3.73 \\
\hline
\multicolumn{7}{l}{\textit{Questions requiring $k=3$ relevant graphs ($n=117$)}} \\
1$^{\dagger}$ & 2.38 & 2.59 & 3.11 & --- & --- & --- \\
3$^{\dagger}$ (all relevant) & 4.32 & 4.38 & 4.40 & --- & --- & --- \\
4 & 4.25 & 4.24 & 4.34 & 4.03 & 4.18 & 4.24 \\
6 & 4.04 & 4.14 & 4.22 & 3.74 & 3.90 & 3.98 \\
\hline
\end{tabular}
\end{table}

Three properties hold in every dimension. First, withholding relevant graphs dominates every other manipulation of the input: reducing the input to a single graph costs $1.48$ GPA on items that require two graphs and $1.94$ on those that require three, roughly five times the largest effect of any padding condition. Second, appending irrelevant graphs degrades performance monotonically, and by nearly the same amount in both item sets: six images cost $0.27$ GPA relative to the exact required set, for $k=2$ and $k=3$ alike. At four images, three of the six relevant-first cells fall inside the decoding-noise floor of $0.10$, whereas every six-image cell lies outside it, so the effect becomes measurable only once the irrelevant graphs outnumber the relevant ones. Third, position matters: interleaving the relevant graphs among the irrelevant ones instead of placing them first costs a further $0.15$ GPA at six images for $k=2$ and $0.30$ for $k=3$. Part of what the padding measures is therefore the difficulty of locating the relevant graphs, and not only the cost of holding more images in context.

The three dimensions are ordered consistently by sensitivity. Padding to six images costs $0.27$ GPA in both item sets, but only $0.21$ and $0.25$ IRA and $0.15$ and $0.18$ RCC. Graph parsing degrades first and the internal coherence of the response degrades last: the model continues to produce a well-formed chain of reasoning over whichever graphs it has attended to, which is precisely why GPA, rather than RCC, is the dimension that separates these conditions.

\subsection*{Evaluation Across Graph Group Types}
\label{appendix:graphtype}

\begin{table*}[t]
\centering
\small
\setlength{\tabcolsep}{6pt}
\caption{Performance by graph type under the Qwen3.8-27B judge. ``All'' denotes the overall score, ``Range'' the within-model range across graph types, and ``Avg.'' the average score across models for each graph type. \textbf{Bold} indicates the best result for each graph type within each metric.}
\begin{tabular}{lccccccc}
\toprule
\rowcolor{gray!15}
\multicolumn{8}{c}{\textit{GPA}} \\
\midrule
\textbf{Model} & \textbf{Flow.} & \textbf{Know.} & \textbf{Mind} & \textbf{Route} & \textbf{Heter.} & \textbf{All} & \textbf{Range} \\
\midrule
GPT-5.4 & 4.14 & 4.06 & 3.63 & \textbf{3.42} & 3.90 & 3.84 & 0.72 \\
Gemini-3.1-Pro & \textbf{4.36} & 4.28 & \textbf{4.16} & 2.92 & \textbf{4.50} & \textbf{4.12} & 1.58 \\
Claude-Sonnet-4-6 & 4.12 & 4.30 & 3.94 & 3.18 & 4.32 & 4.03 & 1.14 \\
Qwen3.5-122B-A10B & 4.02 & 4.40 & 3.90 & 3.06 & 4.38 & 4.02 & 1.34 \\
Qwen3.5-35B-A3B & 4.18 & 4.36 & 3.63 & 3.14 & 4.25 & 3.97 & 1.22 \\
Qwen3.5-27B & 4.16 & \textbf{4.48} & 3.84 & 3.24 & 4.33 & 4.06 & 1.24 \\
\midrule
\textbf{Avg.} & 4.16 & 4.31 & 3.85 & 3.16 & 4.28 & --- & --- \\
\midrule
\rowcolor{gray!15}
\multicolumn{8}{c}{\textit{IRA}} \\
\midrule
\textbf{Model} & \textbf{Flow.} & \textbf{Know.} & \textbf{Mind} & \textbf{Route} & \textbf{Heter.} & \textbf{All} & \textbf{Range} \\
\midrule
GPT-5.4 & \textbf{4.36} & 4.32 & 3.80 & \textbf{3.86} & 4.45 & 4.21 & 0.65 \\
Gemini-3.1-Pro & \textbf{4.36} & 4.36 & \textbf{4.04} & 3.34 & \textbf{4.61} & 4.22 & 1.27 \\
Claude-Sonnet-4-6 & 4.34 & \textbf{4.38} & 4.02 & 3.72 & 4.60 & \textbf{4.28} & 0.88 \\
Qwen3.5-122B-A10B & 4.00 & 4.20 & 3.76 & 3.42 & 4.48 & 4.06 & 1.06 \\
Qwen3.5-35B-A3B & 4.22 & 4.34 & 3.53 & 3.58 & 4.44 & 4.09 & 0.91 \\
Qwen3.5-27B & 4.12 & 4.28 & 3.84 & 3.76 & 4.46 & 4.15 & 0.70 \\
\midrule
\textbf{Avg.} & 4.23 & 4.31 & 3.83 & 3.61 & 4.51 & --- & --- \\
\midrule
\rowcolor{gray!15}
\multicolumn{8}{c}{\textit{RCC}} \\
\midrule
\textbf{Model} & \textbf{Flow.} & \textbf{Know.} & \textbf{Mind} & \textbf{Route} & \textbf{Heter.} & \textbf{All} & \textbf{Range} \\
\midrule
GPT-5.4 & 4.26 & 4.36 & 3.80 & \textbf{3.56} & 4.35 & 4.11 & 0.80 \\
Gemini-3.1-Pro & \textbf{4.36} & 4.34 & \textbf{4.29} & 3.14 & \textbf{4.58} & 4.22 & 1.44 \\
Claude-Sonnet-4-6 & 4.28 & 4.40 & 4.04 & 3.42 & 4.52 & 4.20 & 1.10 \\
Qwen3.5-122B-A10B & 4.06 & 4.24 & 3.92 & 3.28 & 4.49 & 4.08 & 1.21 \\
Qwen3.5-35B-A3B & 4.18 & 4.30 & 3.76 & 3.36 & 4.43 & 4.08 & 1.07 \\
Qwen3.5-27B & 4.18 & \textbf{4.42} & 3.96 & 3.48 & 4.44 & 4.15 & 0.96 \\
\midrule
\textbf{Avg.} & 4.22 & 4.34 & 3.96 & 3.37 & 4.47 & --- & --- \\
\bottomrule
\end{tabular}
\label{tab:graphtype_all_metrics}
\end{table*}

Table~\ref{tab:graphtype_all_metrics} reports GPA, IRA, and RCC for all six evaluated models across the five graph types. Because the test set is balanced by construction, the per-type scores are directly comparable.

Route maps and mind maps are the two lowest-scoring types in all eighteen model--dimension combinations. Route maps rank last for all six models under GPA and RCC, and for four of the six under IRA; in the two exceptions, GPT-5.4 and Qwen3.5-35B-A3B, mind maps fall below route maps by $0.06$ and $0.05$, both of which lie inside the decoding-noise floor. Heterogeneous groups are the highest-scoring type for all six models under IRA and for five of the six under RCC. The difficulty ordering of the graph types is therefore a property of the visual format rather than of the model.

The dominance of graph type over model identity likewise holds in all three dimensions. Averaged over models, a model's spread across the five types is $1.21$ GPA, $0.91$ IRA and $1.10$ RCC; averaged over types, the spread across models within a single type is $0.48$, $0.35$ and $0.33$. The ratio between the two is $2.5\times$ for GPA, $2.6\times$ for IRA and $3.3\times$ for RCC. Aggregate scores over a type-balanced test set therefore understate how strongly the remaining headroom is concentrated in specific visual formats. Aggregate rankings can invert within a type: Gemini-3.1-Pro leads on GPA overall and on heterogeneous groups, yet is the weakest of the six on route maps. This is the intended behavior of a benchmark designed to probe graph-structure competence, and it is also why per-type reporting is necessary: a single aggregate score would conceal the dimension along which these models actually differ.

\subsection*{What Text Serializations Fail to Preserve}
\label{appendix:text_cases}

Table~\ref{table:appC_text_cases} isolates four items on which the subject model is correct under both image-based conditions, i.e., with the graph image alone and with the image plus the structured listing. The errors in the two text-only conditions can therefore be attributed to the serialization itself rather than to question difficulty.

The common failure is the loss of visually encoded information, but the two serializations lose different aspects of it. The structured node--edge listing preserves connectivity while discarding rendering-dependent cues such as edge weights, labels, and captioned annotations. The verbatim transcription retains more local text, but it does not support path tracing, spatial reasoning, or comparison of global layouts. This difference is systematic rather than anecdotal. Although the two text-only conditions have nearly identical mean GPA ($3.71$ for verbatim transcription and $3.72$ for structured listing; paired $p=0.991$), they differ by at least one GPA point on $43.0\%$ of the 300 test items, with 66 favoring the transcription and 63 favoring the listing. Their similar averages therefore mask two largely distinct failure sets.

The four examples make this distinction concrete. In \emph{Know.\ graph / c152}, the question asks for the weight printed on an arrow. The structured listing preserves the edge but omits its label, whereas the verbatim transcription retains the number and enables a correct answer. In \emph{Mind map / c146}, the listing collapses a captioned arrow into an unlabeled relation, causing the model to miss one CO$_2$ source; the transcription preserves the caption and succeeds. In \emph{Route map / c7}, the highway identifier is written along the road, and both serializations fail because the map is reduced to adjacency without the visual cue needed to identify the road itself. In \emph{Heterogeneous / c41}, the task requires comparing hierarchy and complexity across three graphs, which depends on whole-figure organization that neither text representation preserves.

\begin{figure}[h]
\centering
  \includegraphics[width=\linewidth]{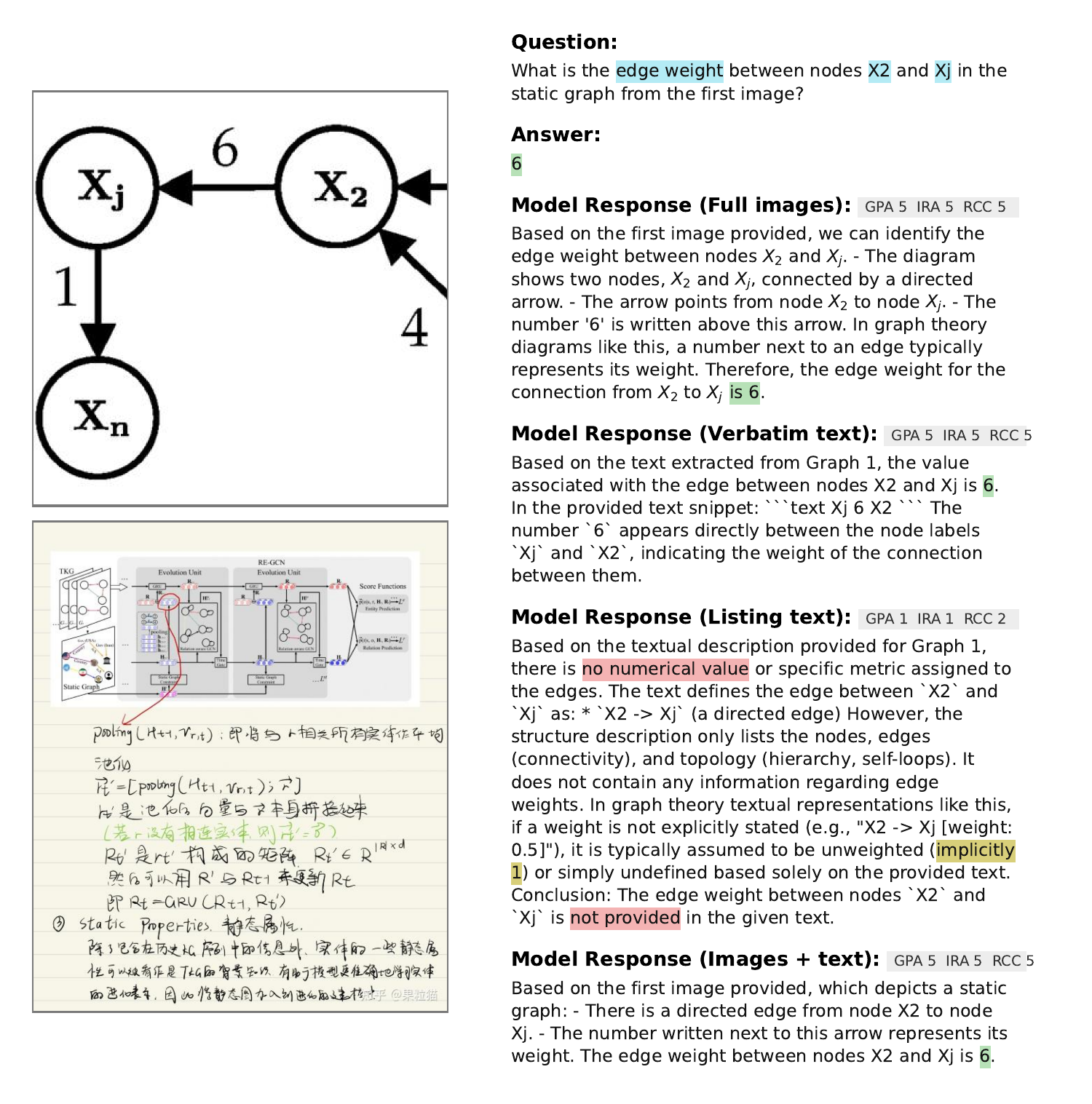}
  \caption{Qualitative results of the text-input ablation on \emph{Know.\ graph / c152}. {\color{red}\text{Red}} denotes incorrect answers, {\color{green}\text{green}} denotes correct answers, and {\color{blue}\text{blue}} denotes the key information in the question.}
  \label{fig:ablation_example1}
\end{figure}

\begin{figure}[h]
    \centering
  \includegraphics[width=\linewidth]{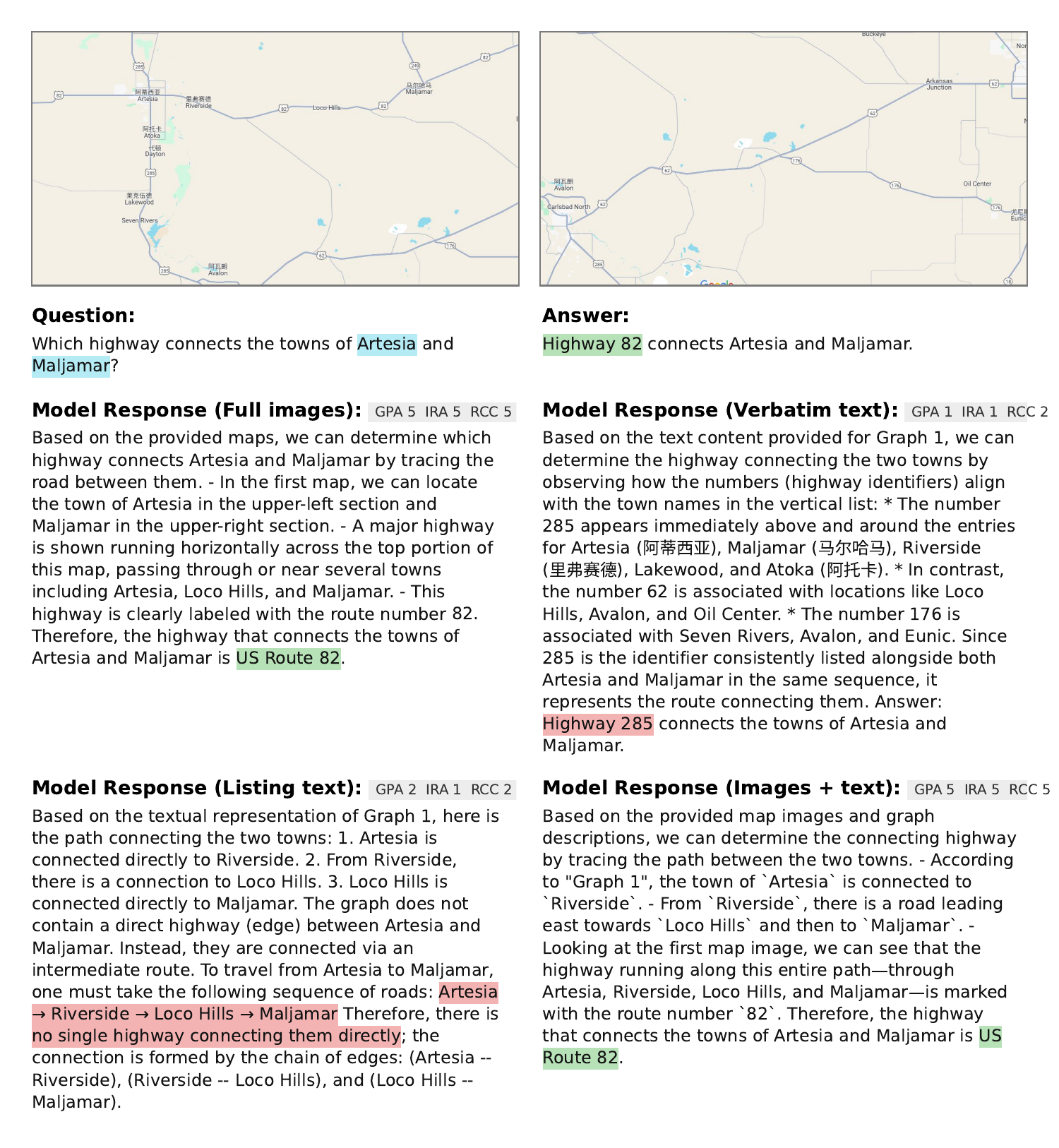}
  \caption{Qualitative results of the text-input ablation on \emph{Route map / c7}. {\color{red}\text{Red}} denotes incorrect answers, {\color{green}\text{green}} denotes correct answers, and {\color{blue}\text{blue}} denotes the key information in the question.}
  \label{fig:ablation_example2}
\end{figure}

These cases illustrate the broader ablation result. Figure~\ref{fig:ablation_example1} shows \emph{Know.\ graph / c152}, where the structured listing loses the edge weight, and Figure~\ref{fig:ablation_example2} shows \emph{Route map / c7}, where both text serializations fail to preserve the road label needed for the answer. More broadly, the two text pipelines are not interchangeable, but neither covers the full set of cues required by the task. The image does. This is why adding structured text to the image yields no gain, whereas adding the image to the same text improves performance by $0.35$ GPA.

\begin{table*}[t]
  \caption{Per-item GPA/IRA/RCC on the 1--5 scale for four illustrative test items, under the Qwen3.8-27B judge; $n_g$ is the number of graphs in the item. Two columns are text-only baselines and two are image-based conditions.}
  \label{table:appC_text_cases}
  \centering
  \small
  \setlength{\tabcolsep}{4pt}
  \begin{tabular}{llccccc}
\hline
 & & & \textbf{Images} & \textbf{Verbatim} & \textbf{Listing} & \textbf{Images} \\
\textbf{Item} & \textbf{Diff.} & \textbf{$n_g$} & \textbf{only} & \textbf{text} & \textbf{text} & \textbf{$+$ text} \\ \hline
Know. graph / c152 & easy & 2 & 5/5/5 & 5/5/5 & 1/1/2 & 5/5/5 \\
Route map / c7 & easy & 2 & 5/5/5 & 1/1/2 & 2/1/2 & 5/5/5 \\
Mind map / c146 & easy & 3 & 5/5/5 & 5/5/5 & 3/3/3 & 5/5/5 \\
Heterogeneous / c41 & difficult & 3 & 5/5/5 & 2/3/3 & 2/4/3 & 5/5/5 \\
\hline
\end{tabular}
\end{table*}

\subsection*{Which Tasks Require Cross-Graph Reasoning?}
\label{appendix:taxonomy}

To test whether the benchmark measures cross-graph structural reasoning rather than general multi-image comparison, we labeled all 300 test items according to what answering them demands: \emph{cross-graph structural} items require node--edge relations from more than one graph; \emph{text-extraction-only} items require reading strings from several images but no structural relation; \emph{single-graph structural} items require node--edge relations within a single graph, even when that graph appears as part of a multi-graph item; and \emph{ordinary multi-image} items require neither. The resulting distribution is $191$, $45$, $57$ and $7$ items, so $63.7\%$ of the test set demands cross-graph structure and only $2.3\%$ reduces to ordinary multi-image VQA. The taxonomy classifies items by what the answer requires, while the $k$ counts referenced by Figure~8 of the main paper and Table~\ref{table:appC_imgcount} record how many graph images the item natively presents; the two dimensions therefore do not partition the test set the same way.

Table~\ref{table:abl_taxonomy} shows that the ablations separate these categories exactly as the labels predict, which validates both the taxonomy and the metric. Restricting the input to a single graph costs $1.81$ GPA on cross-graph structural items but only $0.94$ on single-graph structural ones. Replacing the images with their figure text is nearly free on text-extraction-only items ($-0.04$) and clearly damaging on cross-graph structural items ($-0.35$); adding structured text to the images helps in no category. Inserting irrelevant graphs at shuffled positions likewise costs $0.51$ GPA on cross-graph items against $0.13$ on single-graph structural ones, since only the former require the model to decide which graphs are relevant. The categories that might be suspected of being OCR in disguise are therefore both a minority of the benchmark and identifiable by their ablation signature.

\begin{table*}[!t]
  \caption{Ablation effects by question requirement under automatic evaluation by the Qwen3.8-27B judge. Rows report $\Delta$GPA relative to the full-input condition, computed using paired comparisons within each column-specific item subset.}
  \label{table:abl_taxonomy}
  \centering
  \small
  \begin{tabular}{lcccc}
\hline
 & \textbf{Cross-graph} & \textbf{Text extraction} & \textbf{Single-graph} & \textbf{Multi-image} \\
\hline
Full-input GPA & 4.17 & 4.18 & 3.68 & 4.14 \\
\hline
\multicolumn{5}{l}{\textit{$\Delta$GPA relative to the full-input condition}} \\
Single graph only & $-1.81$ & $-1.91$ & $-0.94$ & $-1.05$ \\
Figure text only, no image & $-0.35$ & $-0.04$ & $-0.67$ & $-0.43$ \\
Images $+$ structured text & $+0.02$ & $-0.04$ & $-0.07$ & $-0.14$ \\
Irrelevant graphs, six images, shuffled & $-0.51$ & $-0.82$ & $-0.13$ & $-0.29$ \\
No graph input & $-2.72$ & $-2.62$ & $-2.05$ & $-2.14$ \\
\hline
\end{tabular}
\end{table*}

\section{Additional Fine-Tuning Analysis}
\label{appendix:ft_analysis}

\begin{table*}[t]
\caption{Detailed fine-tuning results under the Qwen3.8-27B judge. Seed columns are test-set means for independently trained adapters; $\Delta$ is the mean paired item-level change.}
\label{tab:finetuning_statistics}
\centering
\small
\setlength{\tabcolsep}{3pt}
\begin{tabular}{llrrrrrrl}
\toprule
Model & Metric & Before & Seed 42 & Seed 43 & Seed 44 & $\Delta$ & 95\% CI & $p$ \\
\midrule
InternVL2.5-1B & GPA & 1.853 & 2.160 & 2.110 & 2.123 & $+0.278$ & $[+0.182,+0.374]$ & $<10^{-4}$ \\
 & IRA & 2.327 & 2.633 & 2.613 & 2.623 & $+0.297$ & $[+0.191,+0.403]$ & $<10^{-4}$ \\
 & RCC & 2.030 & 2.173 & 2.173 & 2.137 & $+0.131$ & $[+0.037,+0.225]$ & $0.008$ \\
\midrule
InternVL2.5-1B-MPO & GPA & 1.913 & 2.127 & 2.143 & 2.090 & $+0.207$ & $[+0.111,+0.302]$ & $<10^{-4}$ \\
 & IRA & 2.440 & 2.640 & 2.670 & 2.610 & $+0.200$ & $[+0.095,+0.305]$ & $3.2\times10^{-4}$ \\
 & RCC & 2.167 & 2.177 & 2.197 & 2.127 & $+0.000$ & $[-0.093,+0.093]$ & $0.762$ \\
\midrule
DeepSeek-VL-1.3B-Chat & GPA & 1.243 & 1.743 & 1.697 & 1.673 & $+0.461$ & $[+0.382,+0.540]$ & $<10^{-4}$ \\
 & IRA & 1.507 & 2.207 & 2.187 & 2.183 & $+0.686$ & $[+0.587,+0.784]$ & $<10^{-4}$ \\
 & RCC & 1.400 & 1.777 & 1.757 & 1.750 & $+0.361$ & $[+0.279,+0.443]$ & $<10^{-4}$ \\
\midrule
InternVL2-1B & GPA & 1.663 & 1.973 & 1.977 & 1.950 & $+0.303$ & $[+0.198,+0.409]$ & $<10^{-4}$ \\
 & IRA & 2.127 & 2.500 & 2.520 & 2.507 & $+0.382$ & $[+0.271,+0.493]$ & $<10^{-4}$ \\
 & RCC & 1.793 & 2.010 & 2.020 & 2.000 & $+0.217$ & $[+0.118,+0.315]$ & $<10^{-4}$ \\
\midrule
Janus-1.3B & GPA & 1.570 & 1.777 & 1.733 & 1.700 & $+0.167$ & $[+0.082,+0.252]$ & $2.3\times10^{-4}$ \\
 & IRA & 2.003 & 2.240 & 2.227 & 2.217 & $+0.224$ & $[+0.134,+0.315]$ & $<10^{-4}$ \\
 & RCC & 1.670 & 1.870 & 1.817 & 1.777 & $+0.151$ & $[+0.066,+0.236]$ & $7.3\times10^{-4}$ \\
\midrule
mPLUG-Owl3-1B-241014 & GPA & 1.443 & 1.717 & 1.723 & 1.727 & $+0.279$ & $[+0.202,+0.355]$ & $<10^{-4}$ \\
 & IRA & 1.860 & 2.227 & 2.247 & 2.293 & $+0.396$ & $[+0.309,+0.482]$ & $<10^{-4}$ \\
 & RCC & 1.673 & 1.913 & 1.933 & 1.923 & $+0.250$ & $[+0.180,+0.320]$ & $<10^{-4}$ \\
\bottomrule
\end{tabular}
\end{table*}

Table~\ref{tab:finetuning_statistics} reports all 18 model--dimension comparisons under the Qwen3.8-27B judge. For each backbone, we fine-tuned three adapters using seeds 42, 43, and 44, varying the data-shuffling seed accordingly while keeping all other hyperparameters fixed. Each adapter was then evaluated independently with greedy decoding on the full 300-item test set. Base-model outputs were re-scored in the same judging round to ensure direct comparability.

For seed $s$ and test item $i$, let $a_{is}$ denote the post-fine-tuning score and $b_i$ the corresponding base-model score. The reported post-fine-tuning mean is the average of the three seed-specific test-set means, where each seed-specific mean is computed over the 300 test items. The reported SD is the sample standard deviation across these three seed-level means. We further report the mean item-level change,
\[
\Delta=\frac{1}{300}\sum_{i=1}^{300}\left(\frac{1}{3}\sum_{s=1}^{3} a_{is}-b_i\right),
\]
which averages the per-item improvement after first averaging over seeds. Minor discrepancies of 0.001 may arise from independent rounding.

\begin{figure}[h]
  \centering
  \includegraphics[width=\linewidth]{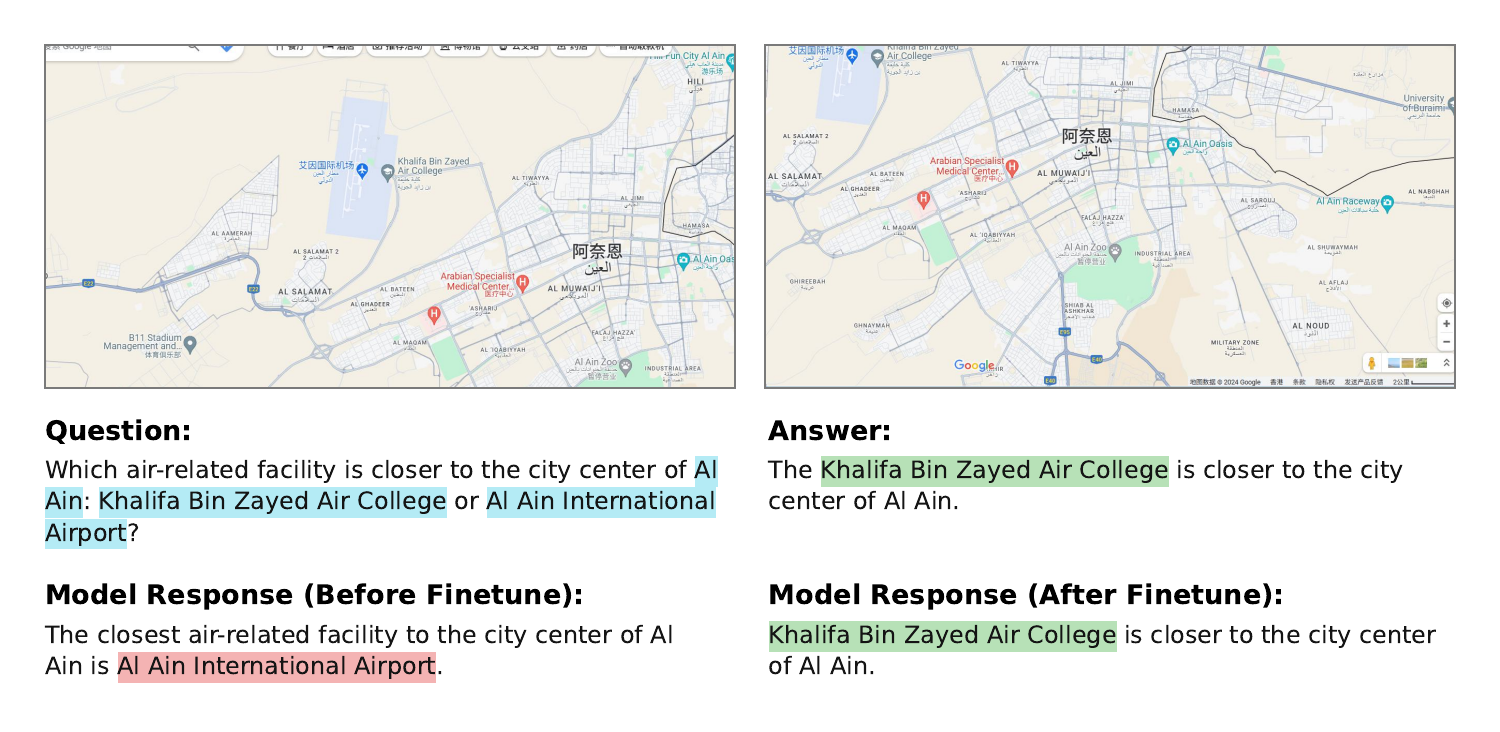}
  \caption{Route Map / c135, InternVL2-1B: the base model names the wrong airport; after fine-tuning the model identifies Khalifa Bin Zayed Air College, matching the reference answer. {\color{red}\text{Red}} denotes incorrect answers, {\color{green}\text{green}} denotes correct answers, and {\color{blue}\text{blue}} denotes the key information in the question.}
  \label{fig:ft_example1}
\end{figure}

The overall pattern is clear. Fine-tuning yields positive mean gains for nearly all model--dimension pairs, with limited variation across seeds: the largest sample SD of the post-fine-tuning test-set mean is 0.047. At the item level, 17 of the 18 comparisons are significant under the paired Wilcoxon signed-rank test ($p<0.05$), including 13 with $p<10^{-4}$. The only exception is RCC for InternVL2.5-1B-MPO ($p=0.762$), whose 95\% normal-approximation interval spans zero. We emphasize that these intervals are computed over test items and should not be interpreted as confidence intervals over training seeds.

\begin{figure*}[h]
  \centering
  \includegraphics[width=0.9\textwidth]{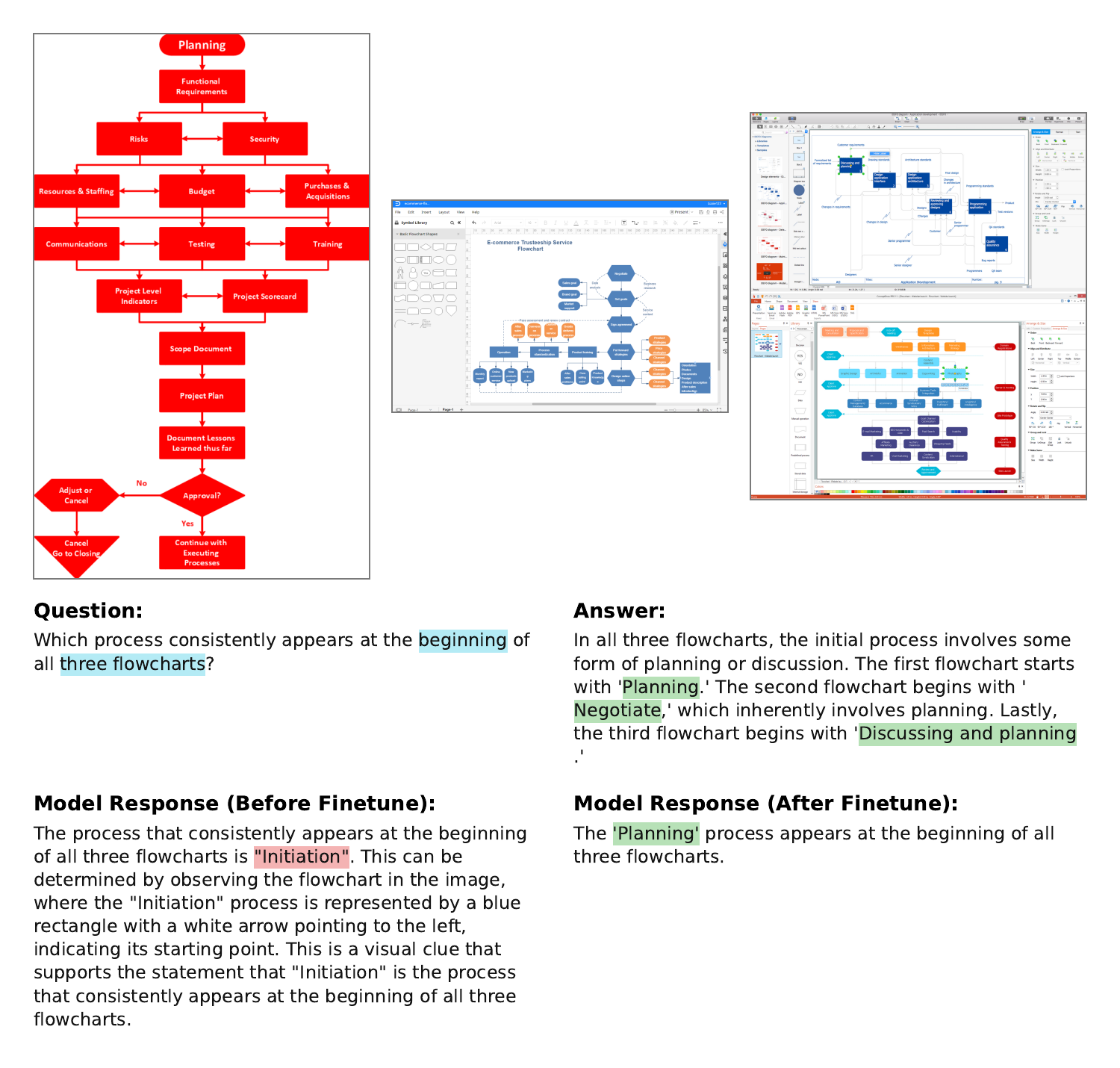}
  \caption{Flowchart / c159, DeepSeek-VL-1.3B-Chat: the base model asserts a fabricated ``Initiation'' step; after fine-tuning the model reports ``Planning'', matching the initial step of the three flowcharts. {\color{red}\text{Red}} denotes incorrect answers, {\color{green}\text{green}} denotes correct answers, and {\color{blue}\text{blue}} denotes the key information in the question.}
  \label{fig:ft_example2}
\end{figure*}

Figures~\ref{fig:ft_example1} and~\ref{fig:ft_example2} provide two representative corrections induced by fine-tuning. In Figure~\ref{fig:ft_example1} (\emph{Route Map / c135}, InternVL2-1B), the base model identifies the wrong airport, whereas the fine-tuned model correctly answers \emph{Khalifa Bin Zayed Air College}. In Figure~\ref{fig:ft_example2} (\emph{Flowchart / c159}, DeepSeek-VL-1.3B-Chat), the base model hallucinates an \emph{Initiation} step, while the fine-tuned model correctly identifies \emph{Planning} as the initial step. These examples illustrate the type of localized reasoning errors that fine-tuning consistently reduces in the aggregate.

\section{Ethical Considerations}
\label{appendix:ethics}
\subsection*{Broader Societal Impacts}
This work introduces a benchmark for MGVR with vision-language models (VLMs), constructed entirely from publicly available or synthetic content to support reproducible academic research.

The benchmark promotes advances in structured visual reasoning and enables fair, systematic evaluation across models, with potential applications in education, science, and healthcare.

While released under the permissive MIT License, we encourage responsible use. Misapplication in high-stakes settings without proper validation may lead to misleading conclusions or unintended outcomes.
Users are advised to carefully assess real-world applicability before deployment.

\subsection*{Dataset Safeguards}
To ensure the safety and ethical integrity of our released dataset, we conducted a manual screening process. While most of the data is synthetically generated or manually created, a small portion was collected from public web sources.

All web-crawled images were individually reviewed by human annotators to ensure they do not contain personal, sensitive, or inappropriate content. This manual screening process ensures the dataset is safe and suitable for academic use under the MIT License.

\end{document}